\documentclass[journal]{IEEEtran}
\usepackage{multirow}
\usepackage{amsmath,graphicx}
\usepackage{algpseudocode}
\usepackage{multirow}
\usepackage{array}
\usepackage{colortbl}
\usepackage{booktabs}
\usepackage{rotating}
\usepackage{caption}
\usepackage{subcaption}
\usepackage[normalem]{ulem} 

\renewcommand\baselinestretch{0.94}

\begin{document}

\title{A ParaBoost Stereoscopic Image Quality Assessment (PBSIQA) System}

\author{Hyunsuk Ko,
        Rui Song,~\IEEEmembership{Member,~IEEE,}
        and~C.-C.~Jay~Kuo,~\IEEEmembership{Fellow,~IEEE}}


\maketitle

\begin{abstract}
The problem of stereoscopic image quality assessment,
which finds applications in 3D visual content delivery such as 3DTV, is
investigated in this work. Specifically, we propose a new ParaBoost
(parallel-boosting) stereoscopic image quality assessment (PBSIQA)
system. The system consists of two stages. In the first stage, various
distortions are classified into a few types, and individual quality
scorers targeting at a specific distortion type are developed. These
scorers offer complementary performance in face of a database consisting
of heterogeneous distortion types.  In the second stage, scores from
multiple quality scorers are fused to achieve the best overall
performance, where the fuser is designed based on the parallel boosting
idea borrowed from machine learning.  Extensive experimental results are
conducted to compare the performance of the proposed PBSIQA system with
those of existing stereo image quality assessment (SIQA) metrics. The
developed quality metric can serve as an objective function to
optimize the performance of a 3D content delivery system.
\end{abstract}


\section{Introduction}

\IEEEPARstart{W}{ith} the rapid development of three-dimensional (3D)
video technology, 3D visual content has become more popular nowadays.
Standards for coding, transmitting and storing 3D visual data have been
proposed such as stereoscopic 3D video~\cite{cit:Stereo}, multiview
video coding (MVC)~\cite{cit:MVC}, and multiview video plus depth map
(MVD) format~\cite{cit:MVD1}. To optimize the performance of a 3D visual
communication system, it is critical to develop a reliable quality
assessment (QA) metric\footnote{The terms "metric", "index" and "score"
are used interchangeably in this paper.} for 3D content.  The PSNR
measure (or any other 2D quality metric) does not correlate well with
human visual experience of 3D visual stimuli~\cite{cit:Lin2014}.  
Since subjective evaluation is time-consuming and
costly, it is desirable to design an objective QA metric that is
consistent with subjective human experience.

While there is a substantial amount of progress in developing 2D
objective image QA indices in recent years, research on 3D image QA is
still preliminary due to several reasons.  The quality of a 2D image is
mainly affected by errors in pixel positions and values.
In contrast, perceptual quality of stereoscopic images
is affected by more factors, including heterogeneous distortions and
mismatch between left and right views, excessive depth perception, etc.
It may result in viewing discomfort such as dizziness and eye strain.
Furthermore, the human visual system (HVS) reacts differently to
asymmetrical distortions caused by different quality levels of left and
right views, depending on distortion types~\cite{cit:Stelmach2000,
cit:Meegan2001, cit:Seuntiens2006}. Another unique stereo distortion is
the rendering distortion, which occurs when the 2D texture image and its
depth data are transmitted simultaneously in the 3D communication
system.  That is, both the texture and the depth data are compressed at
the encoder and transmitted to the decoder and, then, virtual views are
synthesized using the depth-image-based-rendering (DIBR)
technique~\cite{cit:DIBR}. Under this environment, texture errors
trigger blurring or blocking artifacts in virtual views while depth map
errors result in horizontal geometric misalignment since pixel values in
a depth map affect the disparity of the left/right views. 

In order to offer a robust stereo image quality assessment (SIQA) method
against various distortion types, we propose a ParaBoost stereoscopic
image quality assessment (PBSIQA) system in this work.  By ``ParaBoost",
we mean the fusion of multiple image quality scorers into one in order
to boost the overall performance of the quality assessment task.  This
fusion process is achieved by training.  The methodology is also called
``stacking" in the machine learning literature. Although the decision
fusion methodology is generic, the design of individual scorers is
highly application-dependent. To the best of our knowledge, this is the
first work that provides a systematic way to the design of individual
scorers and applies the ParaBoost (or stacking) framework to the
solution of the stereoscopic image quality assessment problem. 

The rest of this paper is organized as follows. Related
previous work is reviewed in Section~\ref{sc:related}.  An overview of
the PBSIQA system is presented in Section~\ref{sc:overview}. Three SIQA
databases are introduced in Section~\ref{sc:database}. The PBSIQA
scoring system is detailed in Section~\ref{sc:learning_model}.
Experimental results and their discussions are provided in
Section~\ref{sc:performance}.  Finally, concluding remarks and future
research directions are given in Section~\ref{sc:conclusions}. 

\section{Review of Previous Work}\label{sc:related}

Traditionally, objective image quality metrics were
classified into pixel-based metrics and the human visual system (HVS)
inspired metrics~\cite{cit:Winkler2005}. More recently, new quality
metrics based on machine learning were proposed. Here, we classify the
design of a quality metric into two approaches: 1) the formula-based
analytical approach and 2) the learning-based prediction approach.  A
quality metric of an analytical form or derived by a parametric model is
constructed using the ``formula-based analytical approach". Both
pixel-based metrics and the human visual system (HVS) inspired metrics
in~\cite{cit:Winkler2005} belong to this class.  A quality metric
derived by a machine learning process (e.g., MMF in \cite{cit:MMF} is
obtained using the ``learning-based prediction approach".

By following the first approach, 2D metrics were extended to 3D metrics
by combining distortions of a depth map and images of both views
linearly in \cite{cit:Campisi2007,cit:Benoit2008}. The
correlation between existing 2D metrics and the perceived quality of a
stereo pair obtained by direct decoding or DIBR was investigated by
Hanhart {\it et al.}~\cite{cit:Hanhart2012, cit:Hanhart2013}.

Another idea is to extract features that influence human visual
experience and combine them to derive a metric.  For example, by
assuming that the perceived distortion and the depth of stereoscopic
images highly depend on local features such as edges and planar parts,
Sazzad {\it et al.}~\cite{cit:Sazzad2012} proposed a SIQA metric for
JPEG compressed images using segmented local features. Some
formula-based metrics exploit HVS characteristics.  Gorley and
Holliman~\cite{cit:Paul2008} proposed a metric based on the sensitivity
of HVS to contrast and luminance changes. Maalou and
Larab~\cite{cit:Maalouf2011} proposed a metric by exploiting the color
disparity tensor and the contrast sensitivity function. Hewage and
Martini~\cite{cit:Hewage2011} presented a reduced-reference quality
metric based on edge detection in the depth map and demonstrated a good
approximation for the full-reference quality metric. Recently, Kim {\it
et al.} \cite{cit:Kim2014} proposed a methodology for subjective 3D QoE
assessment experiments using external stimuli such as vibration,
flickering and sound.

Others use binocular perception models. Stelmach and Meegan
\cite{cit:Stelmach2000, cit:Meegan2001} reported that binocular
perception is dominated by the higher quality image in face of low-pass
filtering operations yet by the average of both images for quantized
distortions. Seuntiens {\it et al.}~\cite{cit:Seuntiens2006} observed
that the perceived quality of a JPEG-coded stereo image pair is close to
the average quality of two individual views. Ryu {\it et
al.}~\cite{cit:RKS2012} proposed an extended version of the SSIM index
based on a binocular model. This metric uses a fixed set of parameters,
but it is not adaptive to asymmetric distortions. Ko {\it et
al.}~\cite{cit:IVMSP} introduced the notion of structural distortion
parameter (SDP), which varies according to distortion types, and
employed the SDP as a control parameter in a binocular perception model
to provide robust QA results for both symmetric and asymmetric
distortions. More recently, Feng {\it et al.} \cite{cit:Feng2013}
proposed an SIQA method by considering the binocular combination
property and the binocular just noticeable difference model.  Ryu {\it
et al.}~\cite{cit:Ryu2014} proposed a no-reference quality metric by
considering perceptual blurriness and blockiness scores and taking
visual saliency into account. 

The second approach has been used to derive 2D and 3D image QA metrics,
and it becomes more and more popular in recent years.  The multi-metric
fusion (MMF) method proposed by Liu {\it et al.}~\cite{cit:MMF} offers
a good example for 2D learning-based image QA, among many others. As to
learning-based SIQA metrics, the number is much less since the size of
stereoscopic image quality databases is relatively small.  Two examples
are given. Park~\textit{et al}.~\cite{cit:Park2012} used a set of
geometric stereo features for anaglyph images and built a regression
model to capture the relationship between these features and the quality
of stereo images.  Cheng and Sumei~\cite{cit:Cheng2012} extracted a set
of basis images using independent component analysis and used a
binary-tree support vector machine (SVM) to predict scores of distorted
stereo images.  

\section{Overview of PBSIQA System}\label{sc:overview}

In this work, we adopt the second approach in the design of a SIQA
system, which is built upon the parallel boosting (ParaBoost) idea.  The
proposed PBSIQA system consists of two stages. At Stage I, multiple
learning-based scorers are designed, where each scorer handles a
specific type of distortion such as the blocking artifact, blurring
distortion, additive noise, and so on. The output of these scorers are a
normilized objective score that has a value between 0 and 1. The output
of these scorers is an obejctive score that only considers the target
distortion. At Stage II, the fuser takes scores from all individual
scorers to yield the final quality score. The prediction model in the
fuser is also obtained from a learning process.  The availability of
multiple scorers at Stage I enables us to handle complex factors that
influence human perceptual quality systematically. In particular, we
will investigate the impacts of various distortions in texture and depth
maps on the quality of rendered images and, then, take them into account
in the design of participating scorers. The design of the PBSIQA system
will be detailed in Section \ref{sc:learning_model}. 

Furthermore, we will show the superiority and the robustness of the
proposed PBSIQA system by extensive experimental results over several
databases consisting of the 2D-plus-depth format as well as the
traditional stereoscopic format with symmetric and asymmetric
distortions in Section \ref{sc:performance}.  Along this line, we
compare the performance of the proposed PBSIQA system against 17 other
2D/3D quality indices.

\section{MCL-3D, IVC-A and LIVE-A Databases}\label{sc:database}

\begin{figure*}[t]
\centering
\begin{subfigure}[h!]{0.15\textwidth}
\centering
\includegraphics[width=1.1in]{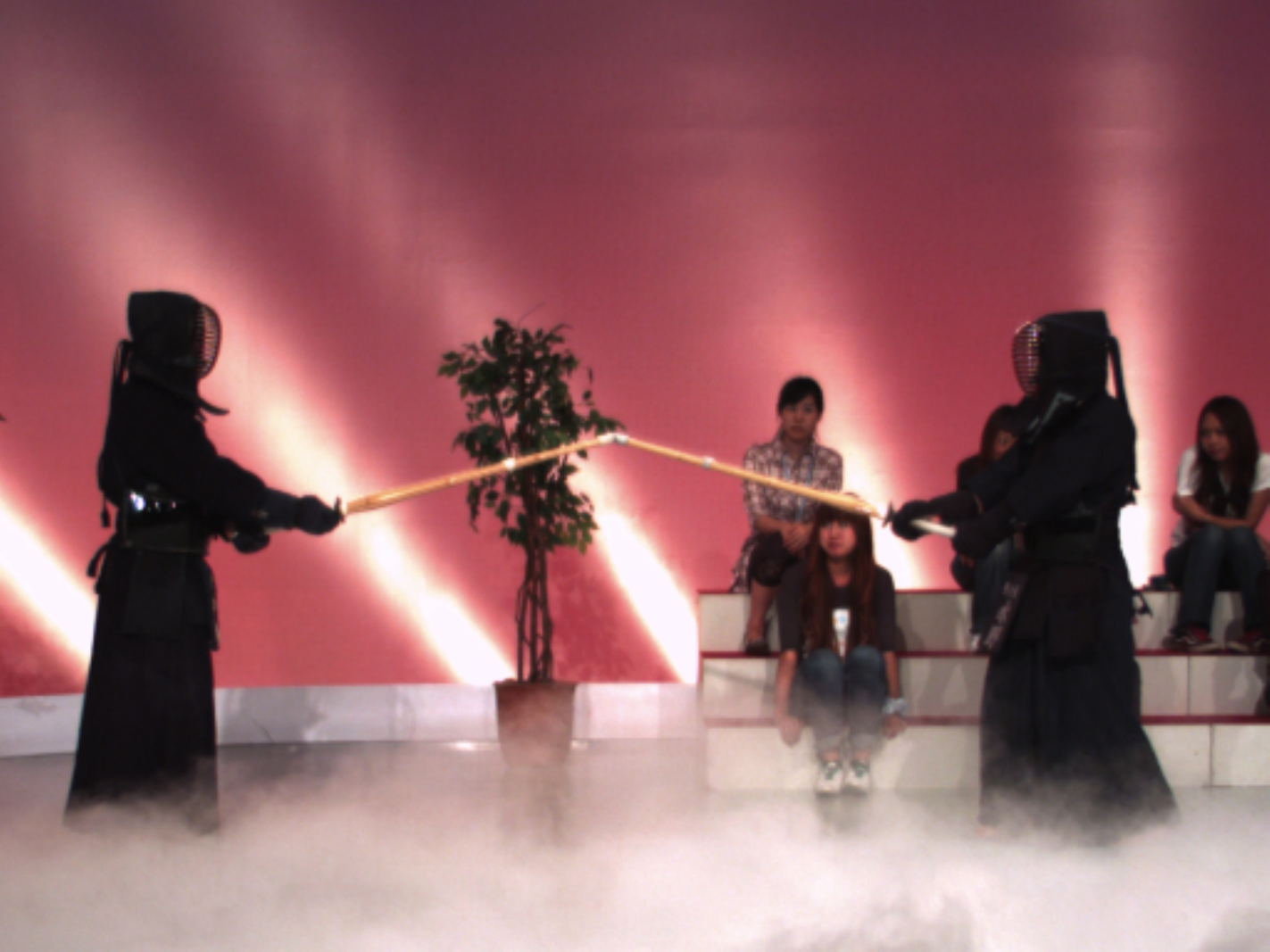} 
\caption{Kendo}\label{fig:kendo}
\end{subfigure}\quad
\begin{subfigure}[h!]{0.15\textwidth}
\centering
\includegraphics[width=1.1in]{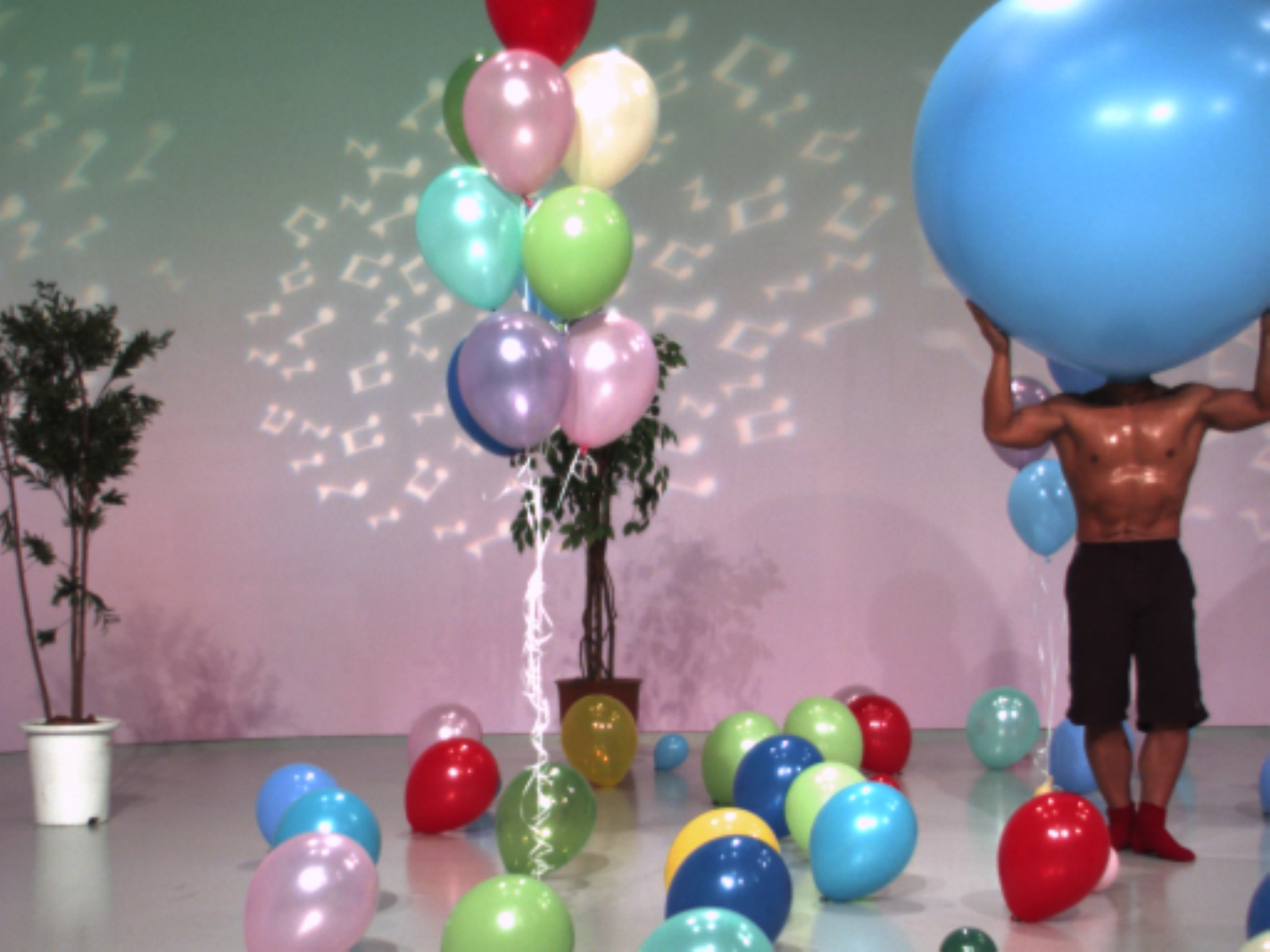} 
\caption{Balloons}\label{fig:balloons}
\end{subfigure}\quad
\begin{subfigure}[h!]{0.15\textwidth}
\centering
\includegraphics[width=1.1in]{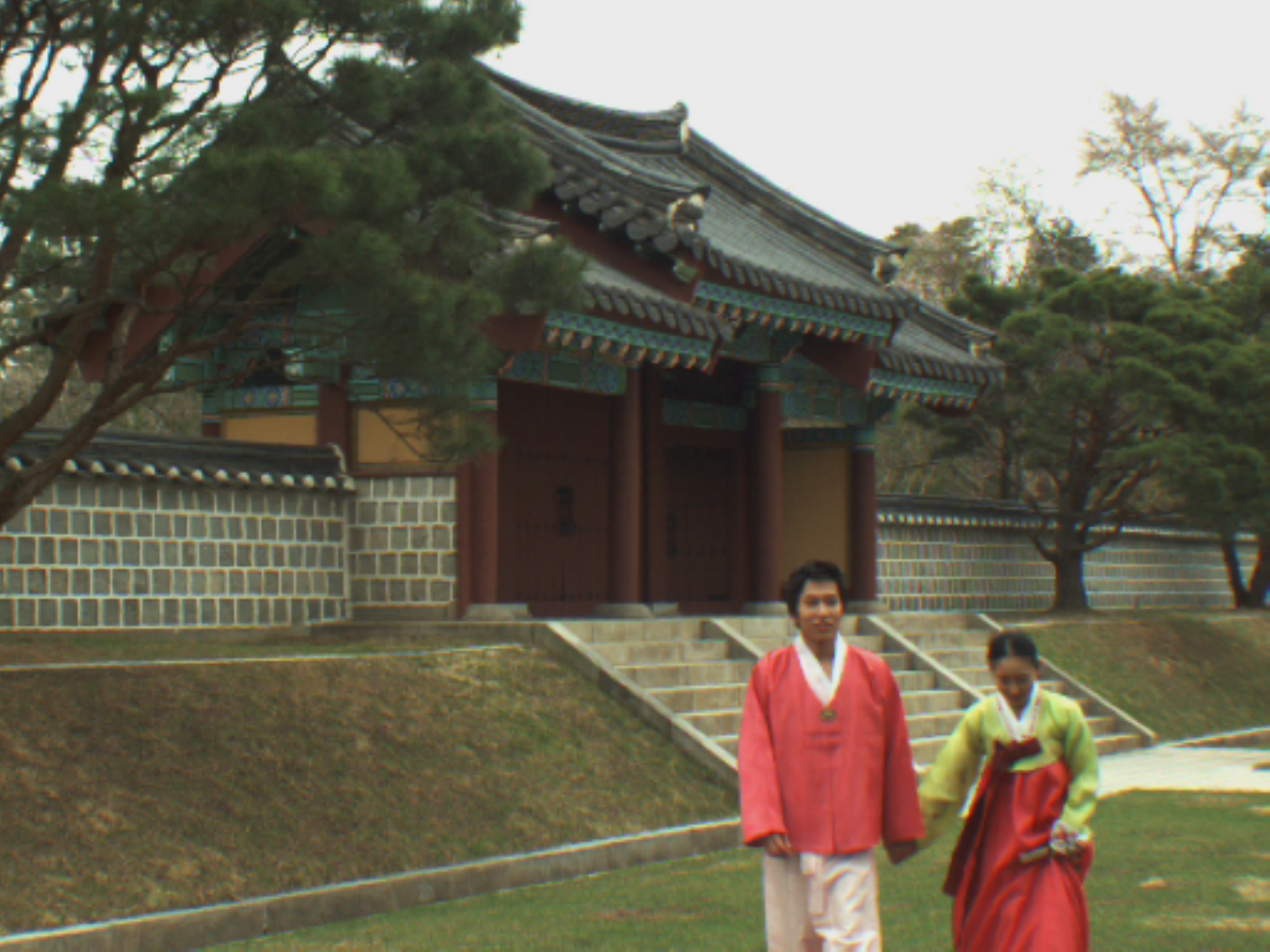} 
\caption{Love\_bird1}\label{fig:love_bird1}
\end{subfigure}\quad
\begin{subfigure}[h!]{0.2\textwidth}
\centering
\includegraphics[width=1.5in]{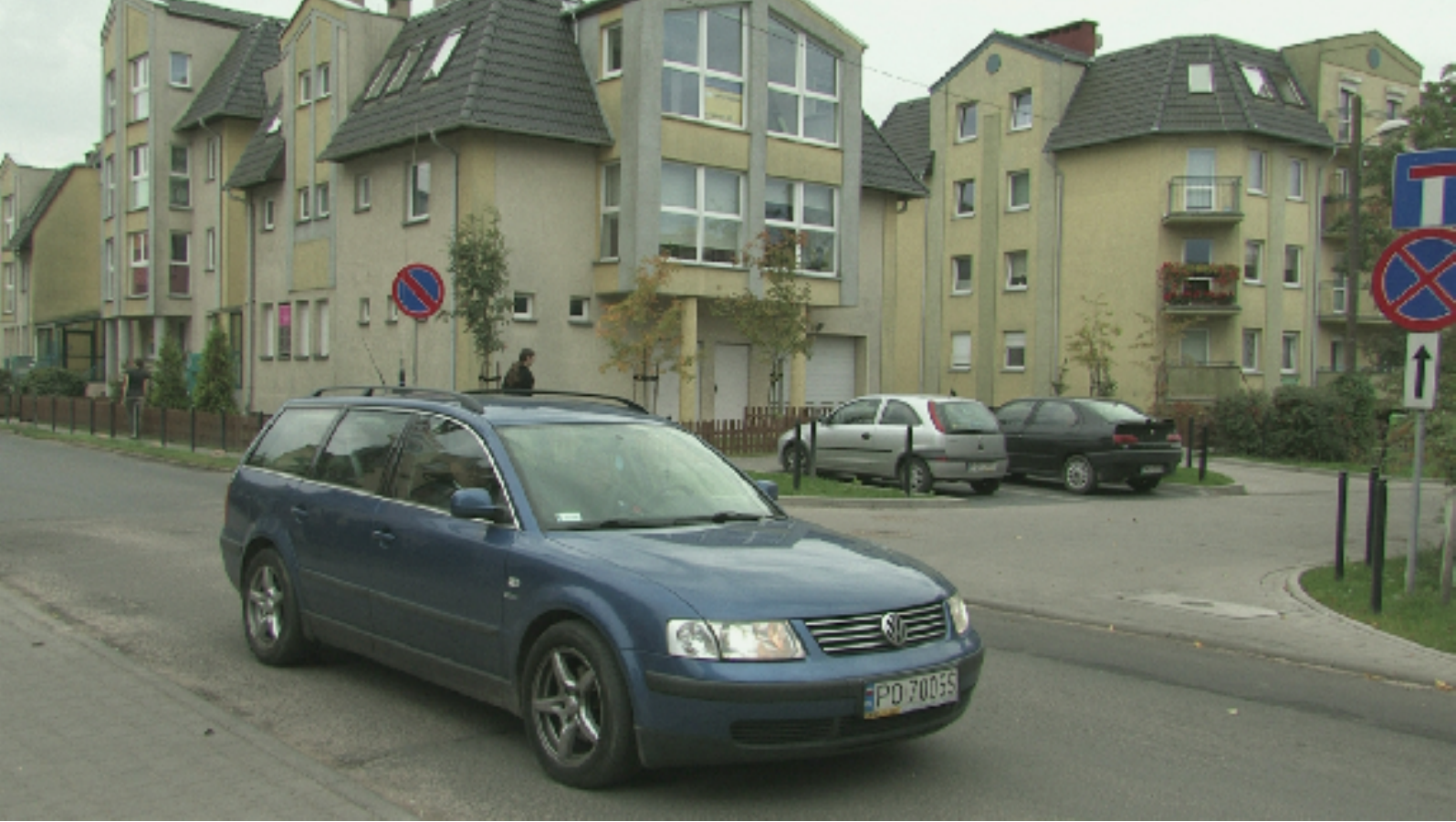} 
\caption{Poznan\_street}\label{fig:WH4poznan_street}
\end{subfigure}\quad
\begin{subfigure}[h!]{0.2\textwidth}
\centering
\includegraphics[width=1.5in]{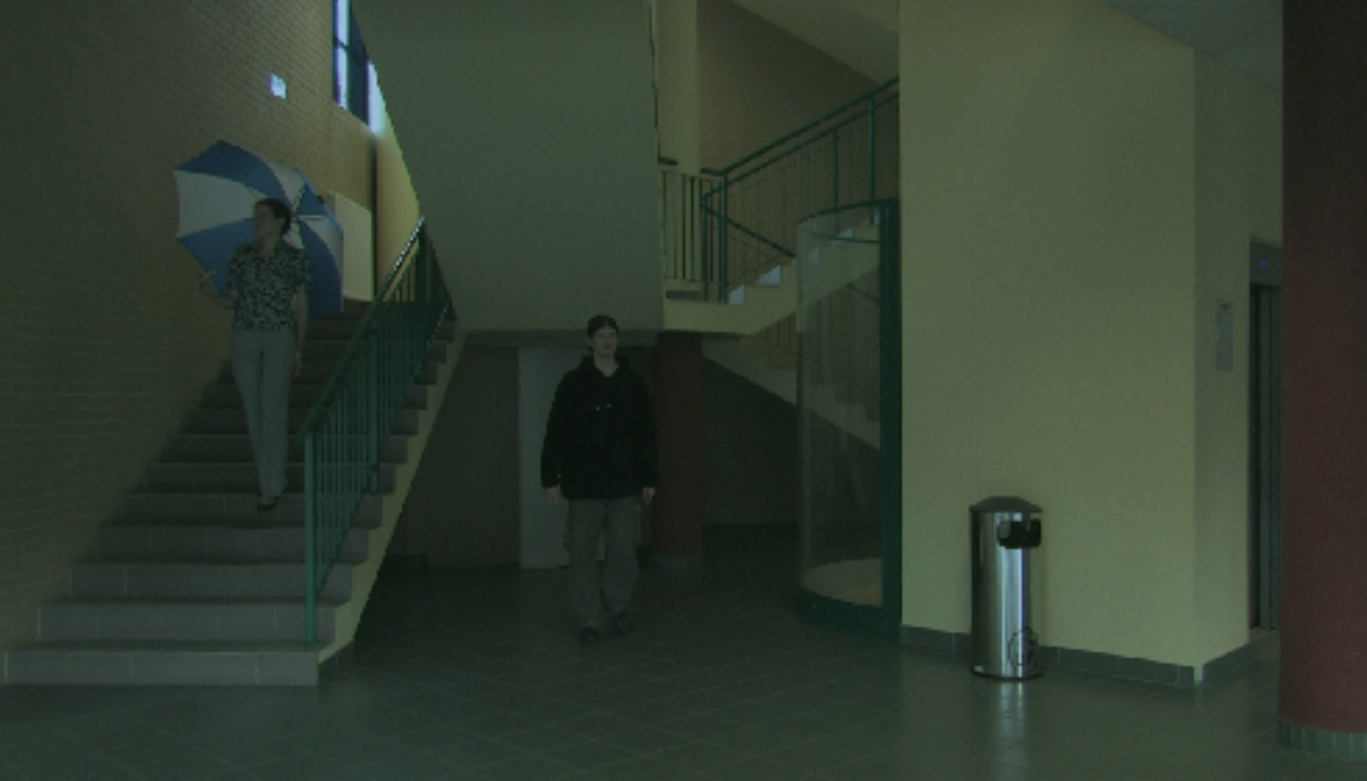} 
\caption{Poznan\_hall2}\label{fig:poznan_hall2}
\end{subfigure}\\
\vspace{0.1in}
\begin{subfigure}[h!]{1.5 in}
\centering
\includegraphics[width=1.5in]{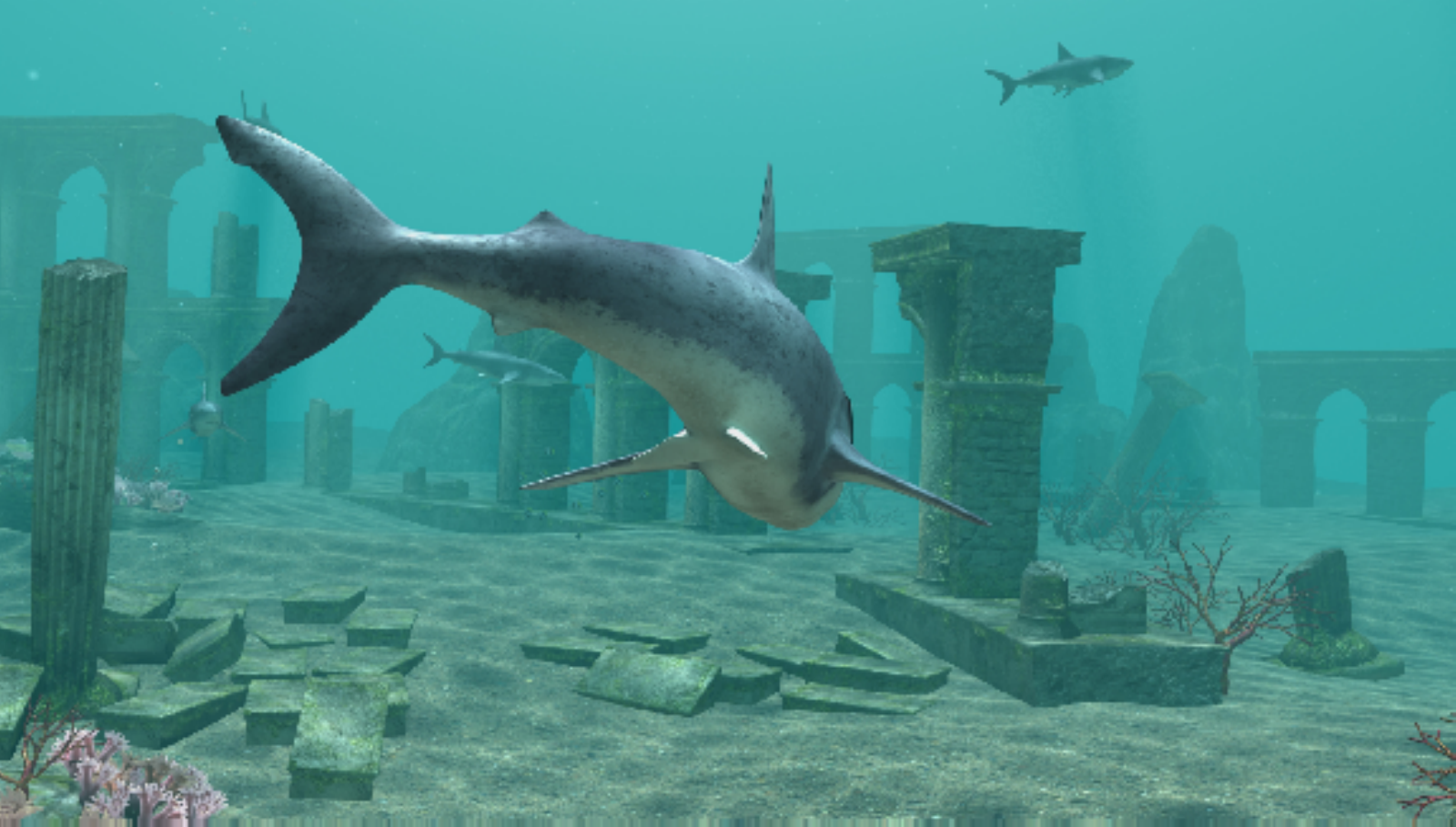} 
\caption{Shark}\label{fig:shark}
\end{subfigure}\quad
\begin{subfigure}[h!]{1.5in}
\centering
\includegraphics[width=1.5in]{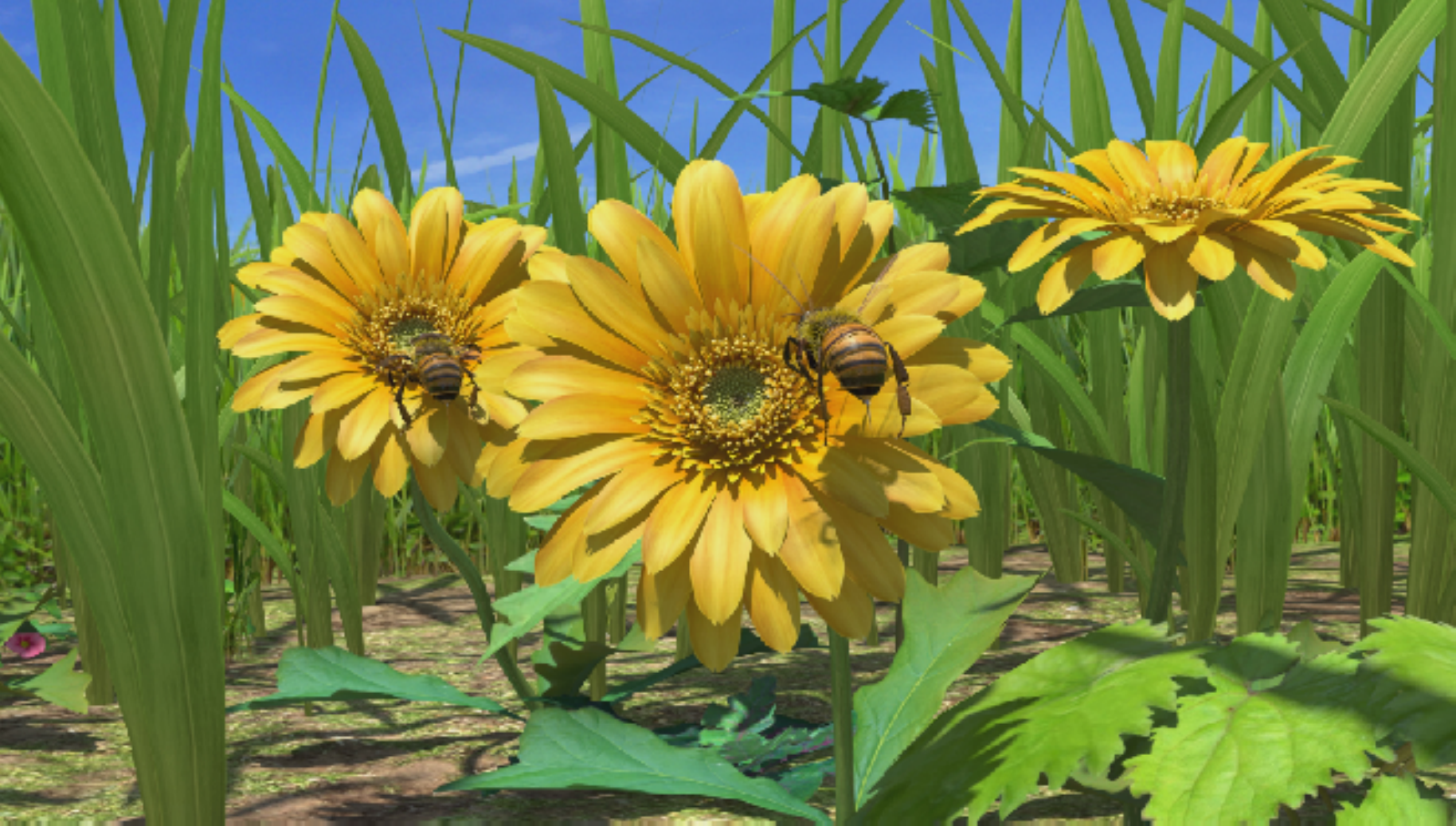} 
\caption{Microworld}\label{fig:microworld}
\end{subfigure}\quad
\begin{subfigure}[h!]{1.5in}
\centering
\includegraphics[width=1.5in]{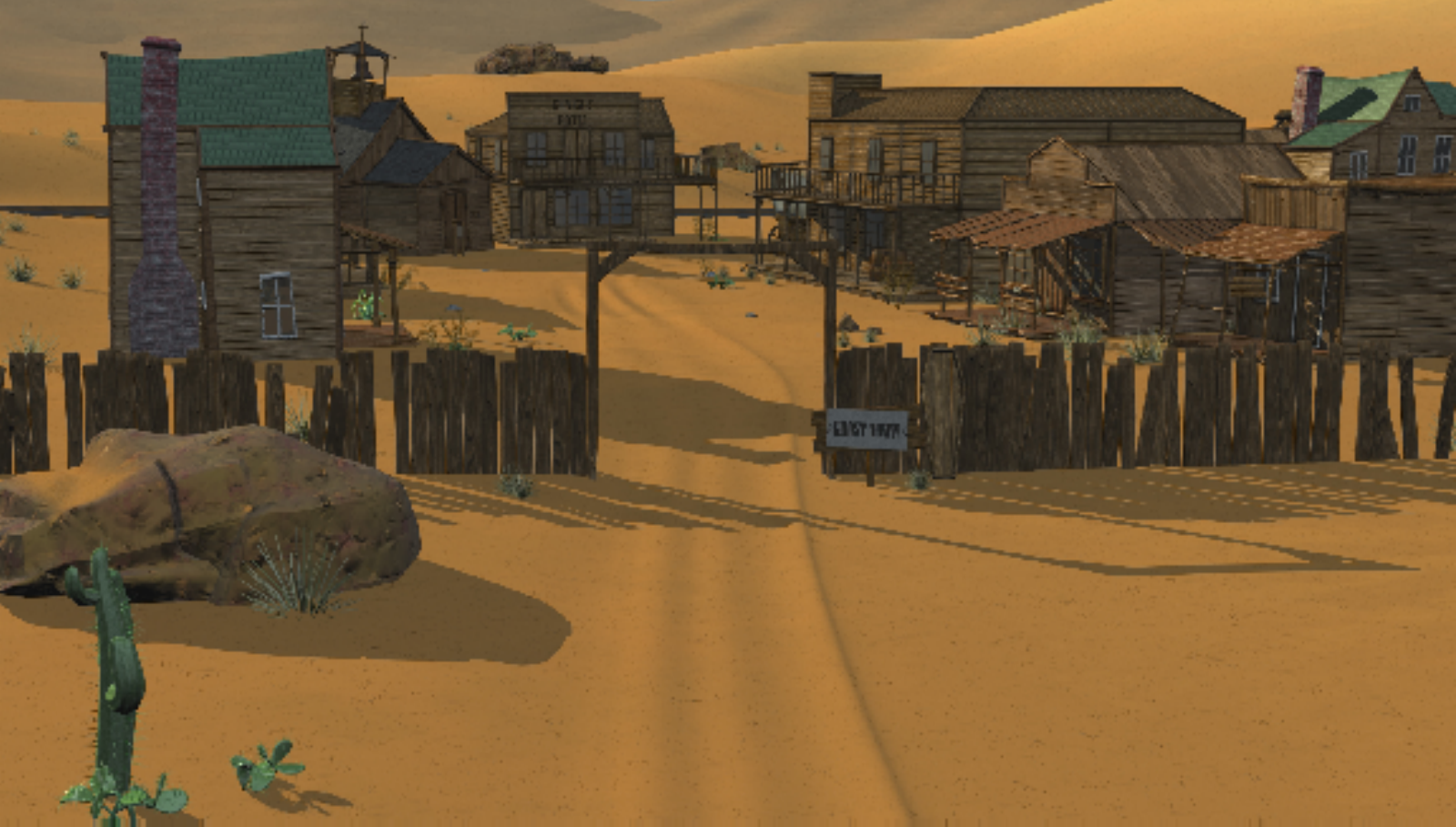} 
\caption{Gt\_fly}\label{fig:gt_fly}
\end{subfigure}\quad
\begin{subfigure}[h!]{1.5in}
\centering
\includegraphics[width=1.5in]{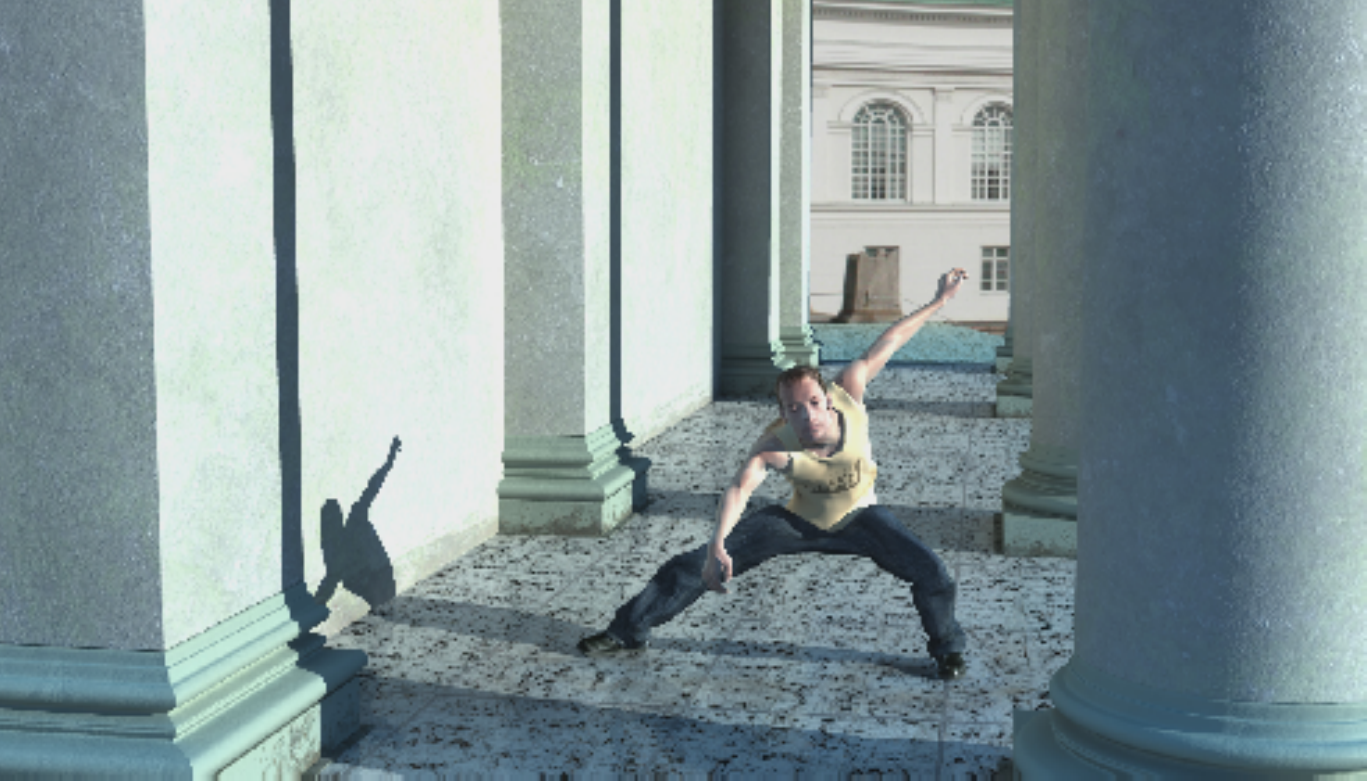} 
\caption{Undo\_dancer}\label{fig:undo_dancer}
\end{subfigure}
\caption{Reference texture images in the MCL-3D database.}\label{fig:ReferenceImage}
\end{figure*}

The 2D-plus-depth format is adopted in the 3DVC standard due to its
several advantages. For example, it allows the texture image and its
corresponding depth map image at a few sampled viewpoints to be
compressed and transmitted in a visual communication system.  After
decoding them at a receiver side, we can render several intermediate
views between any two input views using depth image based rendering
(DIBR) technique. However, very few 3D image quality database was built
based on the 2D-plus-depth format. This is one of the main motivations
in constructing the MCL-3D database~\cite{cit:MCL-3D}. The MCL-3D
database is designed to investigate the impact of texture and/or depth
map distortions on the perceived quality of the rendered image so as to
develop a robust stereoscopic image quality metric.  First, we carefully
chose nine texture images and their associated depth maps from a set of
3DVC test seqeunces as references. 

\begin{figure}[t]
\centering
\includegraphics[width=0.35\textwidth]{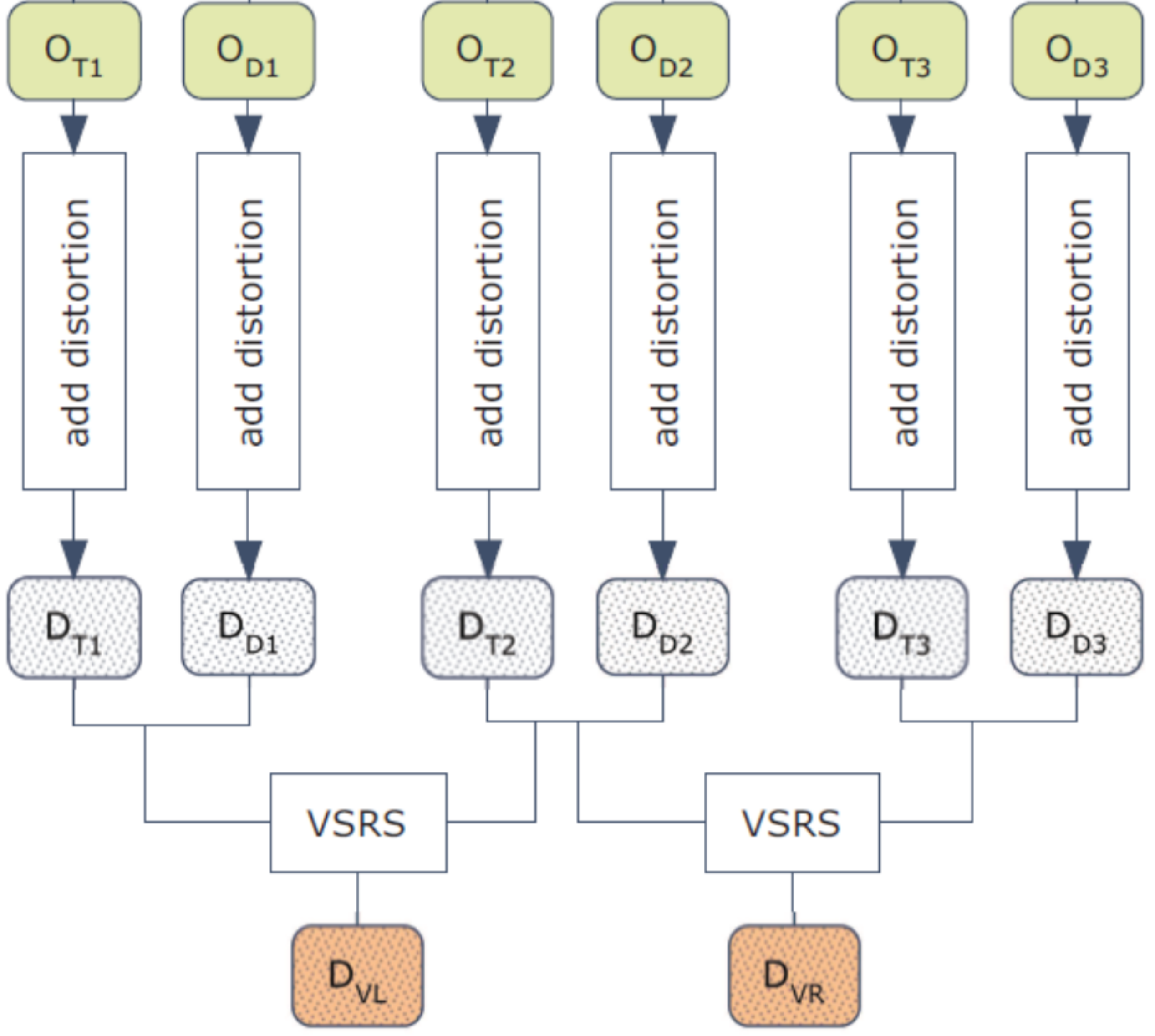}
\caption{Distortion generation in the MCL-3D database.}\label{fig:distortion_design}
\end{figure}

To capture the content variety, the Spatial
Information (SI) defined by ITU-T recommendation~\cite{cit:ITU910} for
each texture image and its depth map was calculated. Also, images were
carefully extracted from an interval with slow motion to avoid motion
blur. This reference set contains indoor/outdoor scenes, CG images,
different depth perception, and so on. In Fig.~\ref{fig:ReferenceImage},
only texture images are shown.  Fig.~\ref{fig:distortion_design} shows
the distortion design.  There are three original views and each view
consists of a texture image and its associated depth map (denoted by
$O_{T1}$/$O_{D1}$, $O_{T2}$/$O_{D2}$, and $O_{T3}$/$O_{D3}$).
Distortions with different types and levels were applied to original
images and/or depth maps. Then, distorted texture images and depth maps
were input to the view synthesis reference software (VSRS 3.5
\cite{cit:VSRS3.5}) along with the configuration file
provided by the production group to render two intermediate views,
$D_{VL}$ and $D_{VR}$ that an assessor will view in the subjective test.
We applied {\em six} different distortions with {\em four} different
levels to three input views symmetrically: Guassian blur, sampling blur,
JPEG compression, JPEG-2000 compression, additive noise, and
transmission error. Based on the recommendations of ITU
\cite{cit:ITU2021}, we consider five quality levels in subjective tests.
The original reference stereoscopic images have excellent quality while
the other 4-level distorted images were controlled by parameters
associated with different distortion types. 

The MCL-3D database consists of the following three sub-databases. 
\begin{itemize}
\item Dataset A contains texture distortions only.
\item Dataset B contains depth map distortions only.
\item Dataset C contains both texture and depth map distortions.
\end{itemize}
The above design allows us to investigate the effect of texture
distortions and depth map distortions independently. 
The test was performed in a controlled environment as recommended by
ITU~\cite{cit:ITU2021}, including the display equipment, viewing
distance, ambient light, etc. We performed pre-screening of subjects and
conducted a training session before the actual test.  The subjective
test results were verified by a screening process according to Annex 2
of Recommendation BT.500~\cite{cit:ITU500}, and outliers were removed.
A pairwise comparison was adopted and the comparison result was
converted to the mean opinion score (MOS). The MCL-3D database can be
downloaded from \cite{cit:MCL-3D}. For further details, we refer to
\cite{cit:Song2015}.

In our earlier work~\cite{cit:IVMSP}, we also constructed two new
databases by expanding the IVC \cite{cit:IVC} and the LIVE
\cite{cit:LIVE} databases and call them IVC-A and LIVE-A, respectively.
We took 12 stereoscopic image pairs from IVC and LIVE as references.
While the distortions in IVC and LIVE are symmetric, we add asymmetric
distortions in IVC-A and LIVE-A. That is, we add distortions of
different levels but the same type to the left and right views of a
stereo image pair. There are four distortion types.  Each stereoscopic
image pair has six combinations of distortion levels: three symmetric
distortion levels ($[1,1], [2,2], [3,3]$) and three asymmetric
distortion levels ($[1,2], [1,3], [2,3]$).  We conducted a subjective
test to obtain MOS using the absolute category rating (ACR)
\cite{cit:ITU2021}.  A summary of the three SIQA databases used in our
test is given in Table \ref{tab:sum_MCLdata}. 

\section{Proposed PBSIQA System}\label{sc:learning_model}

\begin{table*}[t]
  \centering
  \caption{Summary of MCL-3D, IVC-A and LIVE-A Databases.}
    \begin{tabular}{rr|c|c|c}
    \toprule
          &       & \textbf{MCL-3D} & \textbf{IVC-A} & \textbf{LIVE-A} \\
    \toprule
    \multicolumn{1}{c|}{\multirow{12}[0]{*}{\begin{sideways}\textbf{Distortion}\end{sideways}}} & \multicolumn{1}{c|}{\textbf{\shortstack{Reference\\Source}}} & \shortstack{Nine texture/depth map images\\from 3DVC test sequences} & \shortstack{Six stereoscopic image pairs\\from IVC~\cite{cit:IVC} database} & \shortstack{Six stereoscopic image pairs\\from LIVE~\cite{cit:LIVE} database} \\ \cline{2-5}
    \multicolumn{1}{c|}{} & \multicolumn{1}{c|}{\textbf{\# of test images}} & 648   & \multicolumn{2}{c}{144} \\ \cline{2-5}
    \multicolumn{1}{c|}{} & \multicolumn{1}{c|}{\multirow{4}[0]{*}{\textbf{Distortion type}}} & \multirow{4}[0]{*}{\shortstack{6 (gaussian blur / sampling blur /\\JPEG compression / JPEG2000 compression /\\additive noise / transmission error)}} & \multicolumn{2}{c}{\multirow{4}[0]{*}{\shortstack{4 (gaussian blur / JPEG compression /\\ JPEG2000 compression / additive noise)}}} \\
    \multicolumn{1}{c|}{} & \multicolumn{1}{c|}{} &       & \multicolumn{2}{c}{} \\
    \multicolumn{1}{c|}{} & \multicolumn{1}{c|}{} &       & \multicolumn{2}{c}{} \\
    \multicolumn{1}{c|}{} & \multicolumn{1}{c|}{} &       & \multicolumn{2}{c}{} \\ \cline{2-5}
    \multicolumn{1}{c|}{} & \multicolumn{1}{c|}{\textbf{Distortion level}} & 4     & \multicolumn{2}{c}{4} \\ \cline{2-5}
    \multicolumn{1}{c|}{} & \multicolumn{1}{c|}{\textbf{Data format}} & MVD (Multi-View plus Depthmaps) & \multicolumn{2}{c}{Conventional stereoscopic image (only texure image)} \\ \cline{2-5}
    \multicolumn{1}{c|}{} & \multicolumn{1}{c|}{\multirow{6}[0]{*}{\textbf{\shortstack{Distortion\\design}}}} & \multirow{6}[0]{*}{\shortstack{Whole dataset can be divided into three sub-databases.\\
- Dataset A: Distortions only in texture images\\
- Dataset B: Distortions only in depth map images\\
\shortstack{- Dataset C: The same distortion in both texture\\\& depth map images}}} & \multicolumn{2}{c}{\multirow{6}[0]{*}{\shortstack{Each test stereoscopic pair has six combinations\\of distortion per each distortion type as follows:\\-[1,1], [1,2], [1,3], [2,2], [2,3] and [3,3]\\where [l, r] denotes the distortion levels of left and right images\\\& [1]: the strongest, [2]: moderate, and [3]: the weakest distortion.}}} \\
    \multicolumn{1}{c|}{} & \multicolumn{1}{c|}{} &       & \multicolumn{2}{c}{} \\
    \multicolumn{1}{c|}{} & \multicolumn{1}{c|}{} &       & \multicolumn{2}{c}{} \\
    \multicolumn{1}{c|}{} & \multicolumn{1}{c|}{} &       & \multicolumn{2}{c}{} \\
    \multicolumn{1}{c|}{} & \multicolumn{1}{c|}{} &       & \multicolumn{2}{c}{} \\
    \multicolumn{1}{c|}{} & \multicolumn{1}{c|}{} &       & \multicolumn{2}{c}{} \\ \cline{2-5}
    \multicolumn{1}{c|}{} & \multicolumn{1}{c|}{\multirow{2}[0]{*}{\textbf{Remark}}} & \multirow{2}[0]{*}{\shortstack{Rendering distortion\\(two intermediate views are rendered via VSRS3.5)}} & \multicolumn{2}{c}{\multirow{2}[0]{*}{Contain both symmetric and asymmetric distortions}} \\
    \multicolumn{1}{c|}{} & \multicolumn{1}{c|}{} &       & \multicolumn{2}{c}{} \\ \hline
    \multicolumn{1}{c|}{\multirow{4}[0]{*}{\begin{sideways}\textbf{\shortstack{Subjective\\Test}}\end{sideways}}} & \multicolumn{1}{c|}{\textbf{methodology}} & Pair-wise comparison & \multicolumn{2}{c}{\shortstack{ACR-HR\\(w/ 5 grading scales [1:Bad ~ 5:Excellent])}} \\ \cline{2-5}
    \multicolumn{1}{c|}{} & \multicolumn{1}{c|}{\textbf{participant}} & \shortstack{270 people\\(men: 170, women:100 / ages: 21 - 42)} & \multicolumn{2}{c}{\shortstack{20 people\\(men:19, women:1 / ages: 24 - 37)}} \\ \cline{2-5}
    \multicolumn{1}{c|}{} & \multicolumn{1}{c|}{\textbf{Score type}} & MOS   & \multicolumn{2}{c}{MOS} \\ 
    \bottomrule
    \end{tabular}%
  \label{tab:sum_MCLdata}%
\end{table*}%

The impact of texture distortions and depth map distortions on rendering
quality is different through a view synthesis process.  In this section,
we first design scorers tailored to texture distortions and depth map
distortions in Sec. \ref{sec:4.1} and Sec. \ref{sc:scorerdesign_depth},
respectively. The design of these scorers and the score fuser is 
implemented by support vector regression (SVR) as detailed in Sec. 
\ref{sec:4.3}. Finally, the training and test procedures are described 
in Sec. \ref{sc:learning_procedure}.

\subsection{Scorer Design for Texture Distortions}\label{sec:4.1}

\begin{table}[t]
  \scriptsize
  \centering
  \caption{Features used for the scorers}
    \begin{tabular}{>{\centering\arraybackslash}m{1cm}|>{\centering\arraybackslash}m{2cm}|m{0.2cm}|>{\centering\arraybackslash}m{1cm}|>{\centering\arraybackslash}m{1.9cm}|m{0.2cm}}
    \toprule
    \centering \textbf{Type} & \centering \textbf{Feature} & \textbf{Ref} &\textbf{Type} & \centering \textbf{Feature} & \textbf{Ref} \\
    \midrule
    \midrule
    \multirow{7}[17]{*}{\shortstack{Pixel\\difference}} & MD (Maximum Difference) & \cite{cit:Eskicioglu1995} & \multirow{5}[10]{*}{\shortstack{Structural\\Similarity}} & SSIM index & \cite{cit:SSIM}  \\ \cline{2-3} \cline{5-6}
          & MAE (Mean Absolute Error) & \cite{cit:Ismail2002} &       & SSIM Luminance  &  \cite{cit:SSIM} \\ \cline{2-3} \cline{5-6}
          & PSNR  &  -   &       & SSIM Contrast & \cite{cit:SSIM}  \\ \cline{2-3} \cline{5-6}
          & ABV (Average Block Variance) &-       &       & SSIM Similarity & \cite{cit:SSIM}  \\ \cline{2-3} \cline{5-6}
          & MIN (Modified Infinity Norm) & \cite{cit:Ismail2002} &       & UQI (Universal Quality Index) &\cite{cit:Wang2002}  \\ \cline{2-3} \cline{4-6}
          & Blockiness &-  & \multirow{2}[14]{*}{\shortstack{SVD\\related}} & Singular Value & \cite{cit:Narwaria2012} \\ \cline{2-3} \cline{5-6}
          & AADBIIS (Average Absolute Difference Between In-Block Samples) &\cite{cit:MMF}  &       & Singular Vector & \cite{cit:Narwaria2012} \\ \cline{1-3} \cline{4-6}
    \multirow{3}[8]{*}{\shortstack{Edge\\related}} & ES (Edge Stability) &  \cite{cit:Ismail2002} & \multirow{2}[6]{*}{\shortstack{Spectral\\Difference}} & ZCR (Zero Crossing Rate) & \cite{cit:MMF}  \\ \cline{2-3} \cline{5-6} 
          & AES (Average Edge Stability) & \cite{cit:MMF}  &       & PC (Phase Congruency) & \cite{cit:Zhang2011}  \\ \cline{2-3} \cline{3-6}
          & PRATT & \cite{cit:Ismail2002} & \shortstack{Contrast\\measure} & GM (Gradient Magnitude) & \cite{cit:Zhang2011}  \\ \cline{1-3} \cline{4-6}
    \multirow{2}[4]{*}{\shortstack{Image\\Correlation}} & MAS (Mean Angle Similarity) & \cite{cit:Ismail2002} & HVS & HVS-MSE & \cite{cit:Falk2007}  \\ \cline{2-3} \cline{4-6}
          & NCC (Normalized Cross Correlation) & \cite{cit:Ismail2002} & \shortstack{View\\synthesis}   & NDSE (Noticeable Depth Synthesis Error) & \cite{cit:Zhao2011} \\
    \bottomrule
    \end{tabular}%
  \label{tab:candidate_features}%

\end{table}%

In Dataset A of the MCL-3D database, distortions are applied to the
texture image while the depth map is kept untouched. In this case, we
see a similar distortion in the rendered stereo image. For example, 
if the texture image is distorted by a transmission error, a similar distortion 
type is observed in the rendered view since pixel values of the input texture 
image are direct sources to the pixel intensity of the rendered view. 

Furthermore, it is reported in~\cite{cit:IVMSP} that the interaction
between left and right views in perceived 3D image quality depends on
the distortion type.  For example, for blurring and JPEG-2000 coding
distortions, the lost information of a low quality view tends to be
compensated by a high quality view. Thus, the perceptual quality is
closer to that of the high quality view. On the other hand, for additive
noise and JPEG coding distortions, a high quality view is negatively
influenced by a low quality view. For this reason, distortions should be
carefully classified.  We classify distortion types into multiple
groups, and design good scorers for them. For each scorer at Stage I,
 proper features are extracted from input
texture images and trained by a learning algorithm. Then, the
scorer outputs an intermediate score for the target distortion group. 

Based on previous studies on image quality assessment
\cite{cit:Falk2007, cit:Eskicioglu1995, cit:Ismail2002, cit:Zhang2011,
cit:Narwaria2012, cit:Wang2002} and our own experience, we select 24
candidate features for further examination as listed in
Table~\ref{tab:candidate_features}. Then, we calculated
the Pearson correlation coefficient (PCC) to indicate the prediction
performance between MOS and a single-feature-based quality scorer (with
the exception that the singular value and the singular vector are
integrated into one feature vector for the SVD scorer) over Datasets A
and B, respectively. The PCC results of the 23 single-feature-based
scorers are shown in one row of Table~\ref{tab:single_feature_analysis}.
Based on these results, we exclude five features (namely, NCC, AADBIIS,
ABV, AES, and blockiness) whose corresponding scorers have low PCC
values for Datasets A and B. For the remaining ones, we investigate
which features are suitable for which distortion type.  For instance, ES
and Pratt are useful features to be included in the learning-based
scorer targeting at the blurring distortion. In a similar way, we assign
candidate features to several scorers as described below. 

\begin{table}[t]
  \scriptsize
  \centering
  \caption{PCC performance of scorers for Datasets A and B.}
    \begin{tabular}{c|cc|c|cc}
    \toprule
    \multicolumn{6}{c}{\textbf{PCC}} \\
    \midrule
    \textbf{Feature} & \textbf{Dataset A} & \textbf{Dataset B} & \textbf{Feature} & \textbf{Dataset A} & \textbf{Dataset B} \\
    \midrule
    \midrule
    \textbf{ES} & 0.83  & 0.44  & \textbf{HVS\_MSE} & 0.60  & 0.36  \\
    \textbf{MAS} & 0.60  & 0.15  & \textbf{MAE} & 0.68  & 0.63  \\
    \textbf{MD} & 0.73  & 0.63  & \textbf{MIN} & 0.70  & 0.63  \\
    \textbf{NCC} & 0.34  & 0.16  & \textbf{PC} & 0.78  & 0.32  \\
    \textbf{SMPE} & 0.60  & 0.72  & \textbf{PRATT} & 0.78  & 0.56  \\
    \textbf{SSIM index} & 0.64  & 0.67  & \textbf{PSNR} & 0.77  & 0.70  \\
    \textbf{UQI} & 0.68  & 0.45  & \textbf{SSIM\_CON} & 0.68  & 0.68  \\
    \textbf{AADBIIS} & 0.27  & 0.52  & \textbf{SSIM\_LUM} & 0.78  & 0.62  \\
    \textbf{ABV} & -0.20  & 0.38  & \textbf{SSIM\_SIM} & 0.58  & 0.49  \\
    \textbf{AES} & 0.33  & 0.48  & \textbf{ZCR} & 0.64  & 0.17  \\
    \textbf{BLKNESS} & 0.29  & -0.03  & \textbf{SVD} & 0.78  & N/A \\
    \textbf{GM} & 0.78  & 0.69  & \textbf{} &       &  \\
    \bottomrule
    \end{tabular}%
  \label{tab:single_feature_analysis}%
\end{table}%

\noindent{\em Scorer \#1 for Blurring Distortion:} Blurring distortion is
mostly caused by low pass filtering, down-sampling, and compression
(\textit{e.g.} JPEG2000) due to the loss of high frequencies. This kind
of distortion is referred to as the information loss distortion
(ILD)~\cite{cit:RKS2012}. The perceptual quality of a blurred
stereoscopic image pair is closer to that of the high quality view since
the structural component of the high quality view is preserved against
blurring of the low quality view.  Blurring distortion can be observed
around edges most obviously. Human are also sensitive to misalignments
between the edges of left and right views.  Thus, we use two
edge-related features to measure blurriness in an image introduced
below. 

The first one is the edge stability mean squre error (ESMSE)~\cite{cit:Ismail2002}, 
which characterizes the consistency of edges that are evident across
multiple scales between the original and distorted images.  To compute
ESMSE, we first obtain edge maps with five different standard deviation
parameters using the Canny operator.  The output at scale \textit{m} is
decided by thresholding as
\begin{equation} \label{eq:ES1}
E(r,c,\sigma_m) = \left\{ 
  \begin{array}{l l}
    1 & \quad \text{if  } C^m(r,c)>T^m\\
    0 & \quad  otherwise
  \end{array} \right.  
\end{equation}
where $C^m(r,c)$ is the output of the derivative of the Gaussian
operator at the \textit{m}th scale and the threshold
is defined as $T^m=0.1(C_{max}-C_{min})+C_{max}$, where $C_{max}$
and $C_{min}$ are the maximum and minimum values of the norm of the 
gradient output in that band, respectively. An edge stability map $ES(r,c)$ is
obtained by the longest sequence
$E(r,c,\sigma_m),\cdots,E(r,c,\sigma_{m+l-1})$ of edge maps such that
\begin{equation} \label{eq:ES2}
ES(r,c)= l,
\end{equation}
where 
\begin{equation}
\quad l=arg \max \mathop{\cap}_{\sigma_m \leq \sigma_k \leq \sigma_{m+l-1}} E(r,c,\sigma_k)=1.
\end{equation}
The edge stability indices for the original and the distorted images at
pixel location $(r,c)$ are denoted by $Q(r,c)$ and $\hat{Q}(r,c)$,
respectively. Then, the ESMSE value is calculated by summing the edge
stability indices over all edge pixel positions, $n_d$, of the edge
pixels of the original image as
\begin{equation} \label{eq:ES3}
\mbox{ESMSE} = \frac{1}{n_d}\sum_{r,c=0}^{n_d}[Q(r,c)-\hat{Q}(r,c)]^2.
\end{equation}

The other one is Pratt's measure~\cite{cit:Ismail2002} that considers 
the accuracy of detected edge locations, missing edges and false alarm edges. 
The quantity is defined as
\begin{equation}\label{eq:Pratt}
\mbox{Pratt's Measure} = \frac{1}{\max\{n_d,n_t\}}\sum_{i=1}^{n_d}
\frac{1}{1+a \cdot d_i^2},
\end{equation}
where $n_d$ and $n_t$ are the number of the detected and the ground truth 
edge points, respectively, and $d_i$ is the distance to the closest edge for
the $i$th detected edge pixel. The ground truth of the edge image
is obtained from the uncompressed image, and the value of control parameter 
$a$ was set to 0.8 experimentally.\\

\noindent{\em Scorer \#2 for Blocking Distortion:} Blocking distortion
is one of the most common distortions generated by block-based
image/video coders, such as JPEG, H.264 and HEVC. When stereoscopic
images with blocking distortion are shown, the average quality of both
views or that of the lower quality is perceived~\cite{cit:Seuntiens2006,
cit:IVMSP}. The high quality image is negatively influenced by the low
quality image since blocking distortion introduces new visual artifacts
that do not exist in the pristine original scene. Such artifacts are
called the information additive distortion (IAD)~\cite{cit:RKS2012}.  As
stated in \cite{cit:MMF,cit:Falk2007}, there are two good features in
detecting blockiness. They are the HVS-modified MSE (HVS\_MSE) and the
zero crossing rate (ZCR). Mathematically, the HVS\_MSE~\cite{cit:Falk2007} 
is given by
\begin{equation} \label{eq:HVSMSE}
\mbox{HVS\_MSE} = \left\{\frac{1}{RC}\sum_{r=1}^{R}\sum_{c=1}^{C}
|U\{I\}-U\{\hat{I}\}|^2\right\}^{\frac{1}{2}},
\end{equation}
where $R \times C$ is the image size, $I$ and $\hat{I}$ denote original
and distorted images, respectively, $U\{I\}=\mbox{DCT}^{-1}
\{H(\sqrt{u^2+v^2}) \Omega(u,v) \}$ is the inverse discrete cosine
transform (DCT) of $H(\sqrt{u^2+v^2}) \Omega(u,v)$, and where $H(\rho)$,
$\rho=\sqrt{u^2+v^2}$, is an HVS-based band-pass filter and
$\Omega(u,v)$ is the DCT of image $I$. The ZCR along the horizontal
direction is given by
\begin{equation} \label{eq:ZCR1}
\begin{split}
ZCR_h = &\frac{1}{R(C-2)}\sum_{r=1}^{R}\sum_{c=1}^{C-2} z_h(r,c)\\
where \quad z_h(r,c) = &\{
\begin{array}{l l} 
    1, & \quad \text{horizontal ZC at $d_h(r,c)$}\\
    0, & \quad  otherwise
  \end{array}
\end{split}
\end{equation}
and where $d_h(r,c)=\hat{I}(r,c+1)-\hat{I}(r,c)$, $c\in[1,C-1]$. The
vertical zero crossing rate, $\mbox{ZCR}_v$, can be computed similarly. 
Finally, the overall ZCR is given by
\begin{equation} \label{eq:ZCR2}
\mbox{ZCR} = \frac{\mbox{ZCR}_h+\mbox{ZCR}_v}{2}.
\end{equation}
\noindent{\em Scorer \#3 for Additive Noise:} Additive noise is often
caused by thermal noise during a scene acquisition process. If either
view is distorted by additive noise, the perceived quality is degraded
by the low quality image. A pixel-based difference is a good measure for
additive noise.  We select three such features to characterize additive
noise: the peak-signal-to-noise ratio (PSNR), the maximum difference 
(MD)~\cite{cit:Eskicioglu1995} and the modified infinity norm (MIN)
\cite{cit:Ismail2002}. They are defined as
\begin{equation} \label{eq:PSNR}
\mbox{PSNR} = 10 \cdot \log_{10}{\frac{\mbox{255}^2}{\mbox{MSE}}},
\end{equation}
where $\mbox{MSE}=\frac{1}{RC}\sum_{r=1}^{R}\sum_{c=1}^{C}(I(r,c)-\hat{I}(r,c))^2$;
\begin{equation} \label{eq:MD}
\mbox{MD} = \max|I(r,c)-\hat{I}(r,c)|,
\end{equation}
and
\begin{equation} \label{eq:MIN}
\mbox{MIN} = \sqrt{\frac{1}{r}\sum_{i=1}^{r}\bigtriangleup_i^2(I-\hat{I})},
\end{equation}
where $\bigtriangleup_i(I-\hat{I})$ denotes the $i$th largest deviation
among all pixels. In this work, we select the top $25\%$ deviations,
which means $r$ is one fourth of the total number of pixels.\\

\noindent{\em Scorer \#4 for Global Structural Error:} We consider
structural errors since the HVS is highly sensitive to the structural
information of the visual stimuli. Structural errors can be captured by
three components of the SSIM index~\cite{cit:SSIM}, which are luminance,
contrast and structural similarities between images $x$ and $y$. They
are defined as
\begin{eqnarray} \label{eq:SSIMcomp}
L(x,y) &=&\frac{2\mu_{x}\mu_{y}+C_{1}}{\mu_{x}^2+\mu_{y}^2+C_{1}}, \\
C(x,y) &=& \frac{2\sigma_{x}\sigma_{y}+C_{2}}{\sigma_{x}^2+\sigma_{y}^2+C_{2}}, \\
S(x,y) &=& \frac{\sigma_{xy}+C_{3}}{\sigma_{x}\sigma_{y}+C_{3}},
\end{eqnarray}
where $\mu$, $\sigma$, and $\sigma_{xy}$ denote the mean, the standard
deviation and the covariance, respectively. \\
\vspace{-2mm}

\noindent{\em Scorer \#5 for Local Structural Error:} The performance of
an image quality scorer can be improved by applying spatially varying
weights to structural errors. Salient low-level features such as edges
provide important information to scene analysis. Two features were
proposed in~\cite{cit:Zhang2011} at this end: phase congruency (PC) and
gradient magnitude (GM).  Physiological studies show that PC provides a
good measure for the significance of a local structure. Readers are
referred to \cite{cit:Zhang2011} for its detailed definition and
properties. While PC is invariant with respect to image contrast, GM
does take local contrast into account. \\
\vspace{-2mm}

\noindent{\em Scorer \#6 for Object Structure and Luminance Change:} We
utilize singular vectors and singular values as features. Given an image
$I$, we can decompose it into the product of two singular vector
matrices $U$ and $V$ and a diagonal matrix $\Sigma$ in form of
\begin{equation} \label{eq:SVD}
\begin{split}
& U=[{\bf u}_1 {\bf u}_2 \ldots {\bf u}_R], \quad
V=[{\bf v}_1 {\bf v}_2 \ldots {\bf v}_C], \\
& \Sigma = \mbox{diag}(\sigma_1, \sigma_2,...,\sigma_l),
\end{split}
\end{equation}
where ${\bf u}_i$ and ${\bf v}_j$ are column vectors, $\sigma_k$ is a
singular value and $l= \min(R,C)$. It was shown in
\cite{cit:Narwaria2012} that the first several singular vectors offer a
good set to represent the structural information of objects while
subsequent vectors account for finer details. Furthermore, singular
values are related to the luminance changes. They should be considered
since the luminance mismatch between left and right views results in
annoying viewing experience. \\
\vspace{-2mm}

\noindent{\em Scorer \#7 for Transmission Error:} Packet loss and bit
errors occur during stereo image data transmission.  It appreas in form
of block errors since most image coding standards adopt the block-based
approach. Based on the feature analysis, both the universal quality
index (UQI) \cite{cit:Wang2002} and the mean angle similarity (MAS) 
~\cite{cit:Ismail2002} are useful in characterizing transmission error. 
The UQI is defined as
\begin{equation} \label{eq:UQI}
\mbox{UQI} = \frac{\sigma_{xy}}{\sigma_x\sigma_y} \cdot 
\frac{2\mu_x\mu_y}{\mu_x^2+\mu_y^2} \cdot \frac{2\sigma_x\sigma_y}
{\sigma_x^2+\sigma_y^2},
\end{equation}
which is equal to the product of three terms representing the loss of
correlation, the luminance distortion, and the contrast distortion
between two images $x$ and $y$, respectively.  MAS is a feature that
measures the statistical correlation between pixel vectors of the
original and the distorted images since similar colors will result in
vectors pointing to the same direction in the color space. The moments
of the spectral chromatic vector are used to calculte the correlation as
\begin{equation} \label{eq:MAS}
\mbox{MAS} = 1-\frac{1}{N^2}\sum_{r,c=1}^{N}\frac{2}{\pi}\cos^{-1}
\frac{\langle \textbf{C}(r,c), \hat{\textbf{C}}(r,c) \rangle}
{\|\textbf{C}(r,c)\|\|\hat{\textbf{C}}(r,c)\|},
\end{equation}
where $\textbf{C}(r,c)$ and $\hat{\textbf{C}}(r,c)$ indicate the multispectral 
pixel vectors at position $(r,c)$ for original and distorted images, respectively,
and $N$ is the number of pixels that have a non-zero value for 
the inner product and the norm of $\textbf{C}(r,c)$ and $\hat{\textbf{C}}(r,c)$.

For each scorer, three feature values from three input
views, denoted by $V1$, $V2$, and $V3$, are fed into the learning-based
scorers. We entrust the task of seeking an optimal relation among these
features to a learning algorithm, where the impact of the quality
difference between left and right views on the perceptual quality highly
depends on the distortion type. 

\begin{figure}[t]
\centering
\begin{subfigure}[h!]{0.5\textwidth}
\centering
\includegraphics[width=3in]{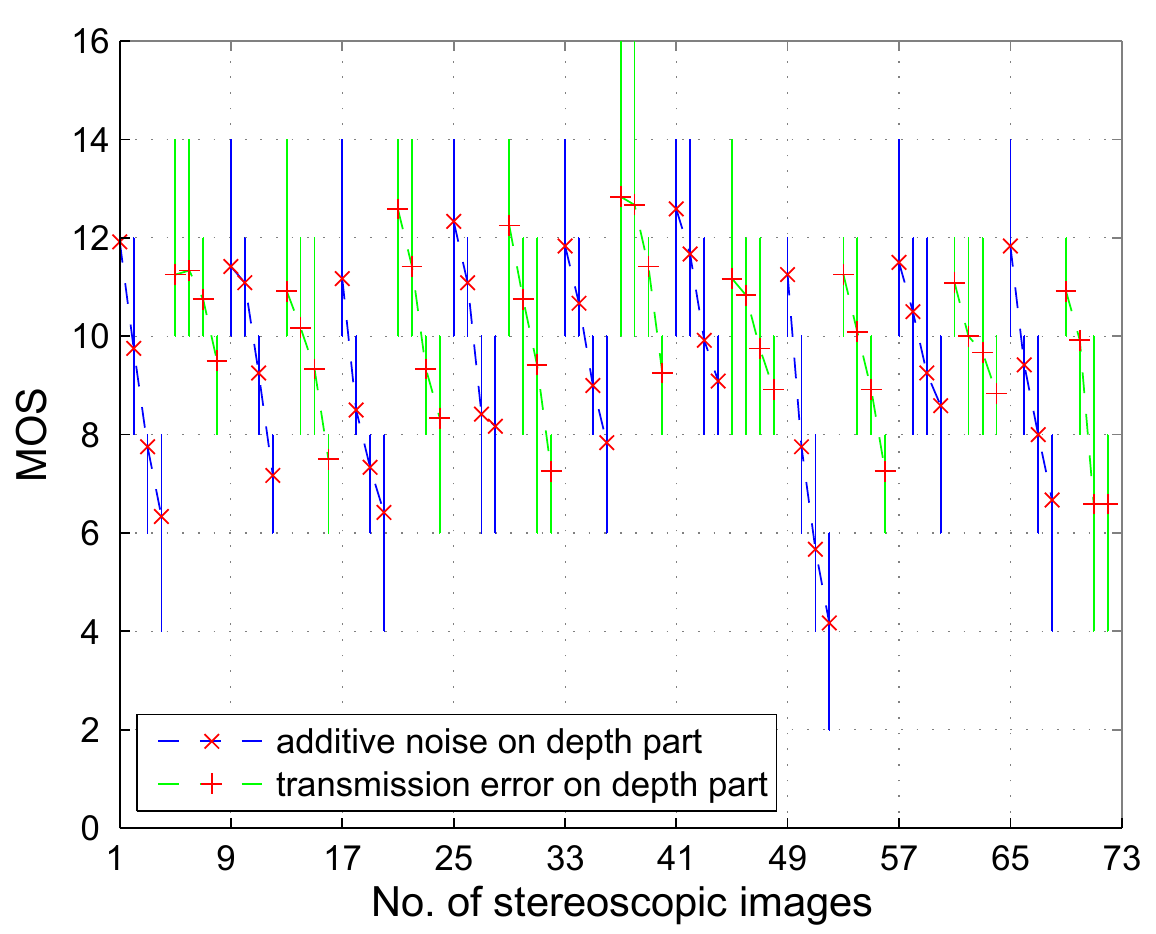}
\caption{The MOS distribution plot with additive noise
and transmission errors in the depth map only.}\label{fig:group1}
\end{subfigure}\\
\vspace{0.1in}
\begin{subfigure}[h!]{0.5\textwidth}
\centering
\includegraphics[width=3in]{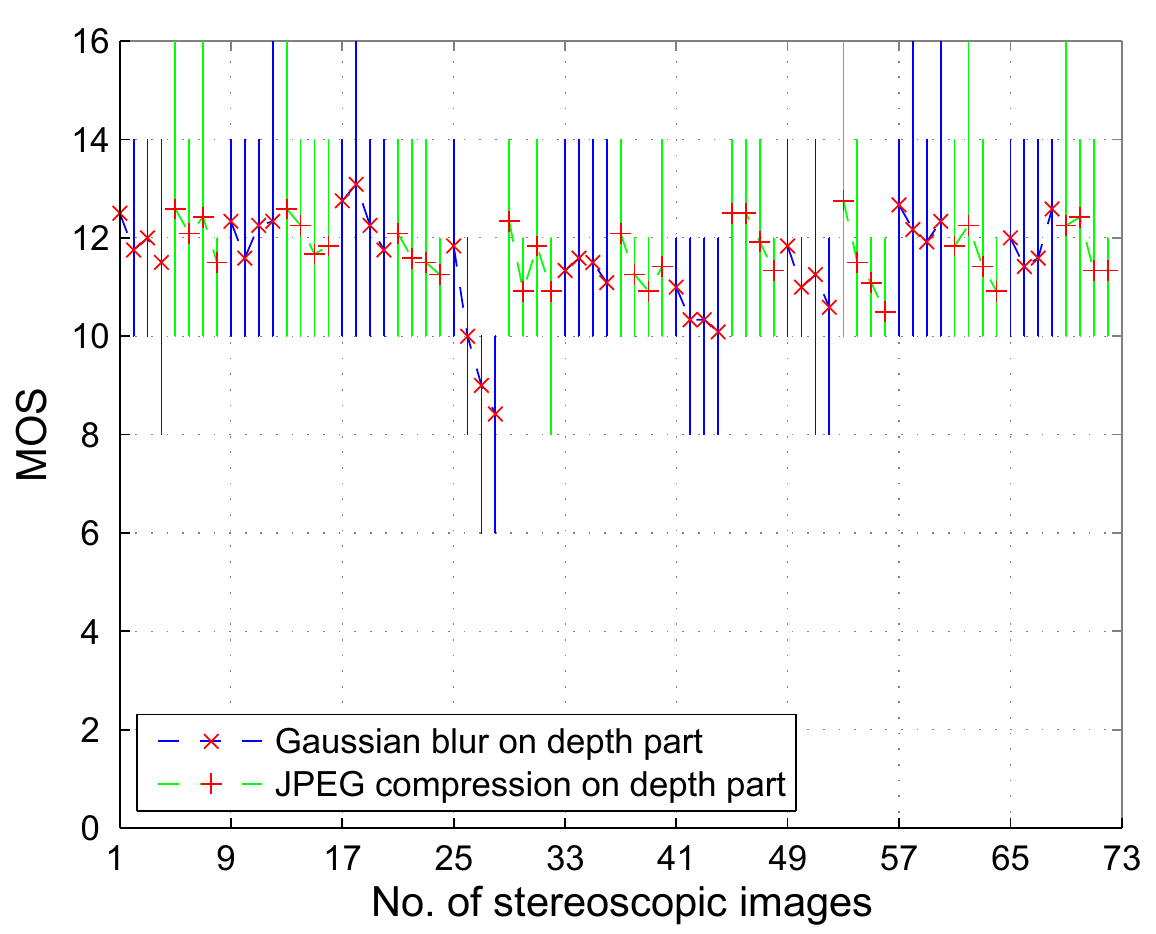}
\caption{The MOS distribution plot with the Gaussian blur 
and the JPEG compression distortion in the depth map only.}\label{fig:group2}
\end{subfigure}
\caption{The plot of the MOS distribution with the 95\%
confidence interval, where the horizontal axis is the image index and
distortions are applied to the depth map only.  Images of the same
source are displayed in the same minor grid. There are two sets of MOS
distribution plots in each minor grid corresponding to two distortion
types, and each set has four plots with increasing distortion
levels.}\label{fig:depth mapMOS}
\vspace{-2mm}
\end{figure}

\subsection{Scorer Design for Depth Distortions}\label{sc:scorerdesign_depth}

Dataset B of the MCL-3D database contains depth map distortions only.
Research on the effect of the depth distortion on rendered stereo image
quality has been conducted recently by quite a few researchers, e.g.,
\cite{cit:Kim2010, cit:Oh2011}. Generally speaking, the depth value is 
inversely proportional to the horizontal disparity of rendered left and right 
views so that the horizontal disparity distortion appears in form of geometric 
errors.

\begin{figure*}[t]
\centering
\includegraphics[width=0.8\textwidth]{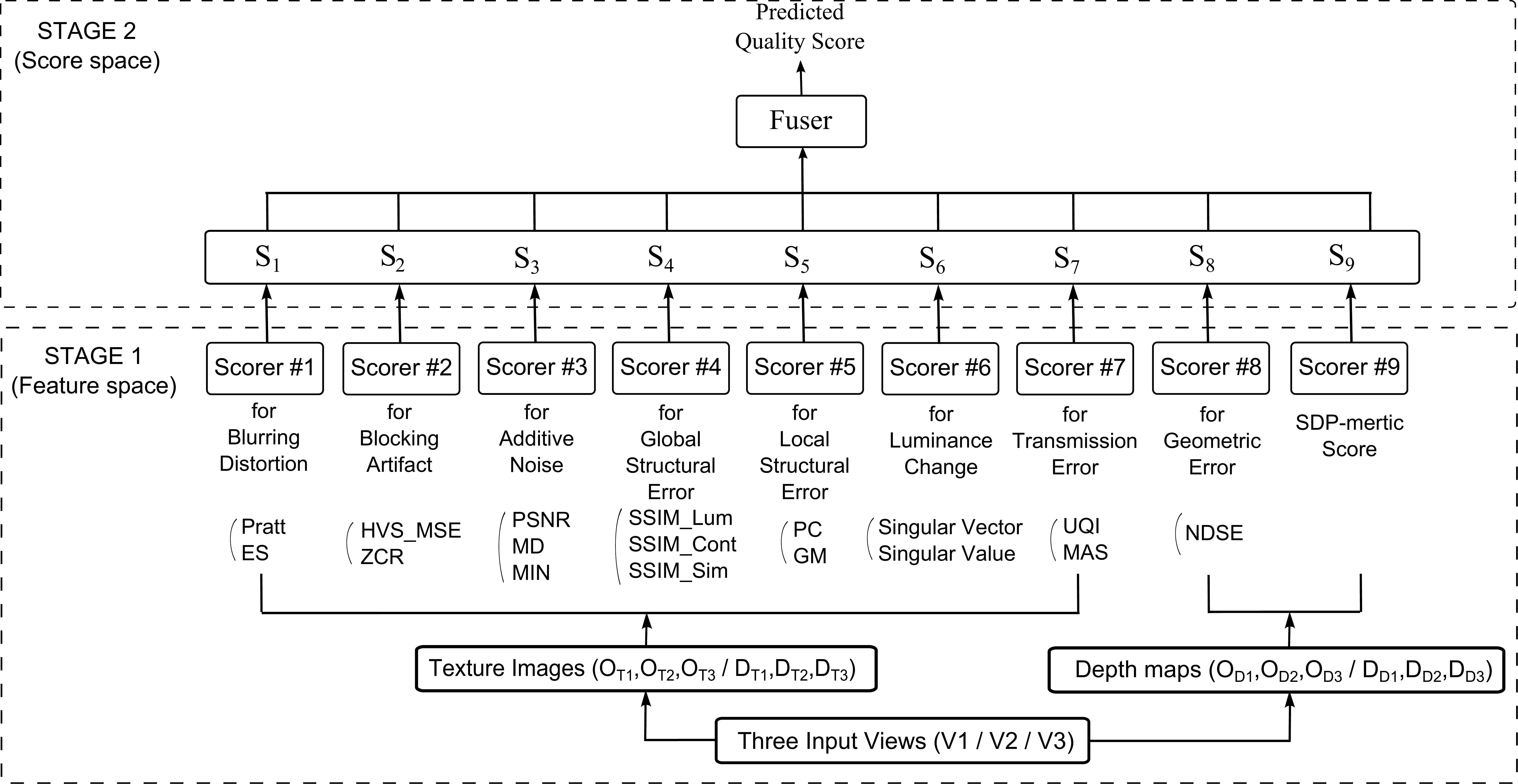}
\caption{The diagram of the proposed PBSIQA system.}
\label{fig:Ensemble_model}
\end{figure*}

\noindent{\em Scorer \#8 for Geometric (Horizontal Disparity) Error:}
Zhao {\it et al.} \cite{cit:Zhao2011} proposed a Depth
No-Synthesis-Error (D-NOSE) model by exploiting that the depth
information is typically stored in 8-bit grayscale format while the
disparity range for a visually comfortable stereo pair is often far less
than 256 levels.  Thus, multiple depth values do correspond to the same
integer (or sub-pixel) disparity value in the view synthesis process. In
other words, some depth distortion may not trigger geometric changes
in the rendered view. Specifically, if a pixel distortion of the
depth map falls into the range defined by the D-NOSE profile, it does not
affect the rendered image quality in terms of MSE. 
Being motivated by the D-NOSE model, we define a noticeable depth
synthesis error (NDSE) feature for geometric errors as
\begin{equation} \label{eq:NDSE}
\mbox{NDSE} = \sum_{i}|D\{i\}-\hat{D}\{i\}|,
\end{equation}
where $D$ and $\hat{D}$ represent the original and the distorted depth maps,
respectively, and $i$ is a pixel index out of the range of D-NOSE profile which 
is defined as
\begin{equation}\label{eq:DNOSE}
\mbox{D-NOSE(v)} = [v+\delta^-(v), v+\delta^+(v)],
\end{equation}
and where 
\begin{eqnarray}
\delta^-(v) & = & \lceil{DP^{-1}(\frac{\lceil{(DP(v)-\lambda})
\cdot K-1\rceil}{K})} \rceil-v, \\
\delta^+(v) & = & \lfloor{DP^{-1}(\frac{\lceil{(DP(v)-\lambda})
\cdot K\rceil}{K})}\rfloor-v,
\end{eqnarray}
$v$ is a quantized depth value, $DP(v)$ is the disparity function of
$v$, $\lambda$ is the offset error for the rounding operation, $K^{-1}$
is the precision, and $\lceil \cdot \rceil$ and $\lfloor \cdot \rfloor$
denote the ceiling and floor operations, respectrively. For more
details on the D-NOSE profile, we refer to \cite{cit:Zhao2011}. 
Since NDSE only considers pixel distortions that change the original
disparity value, it has the highest prediction accuracy (with PCC=0.73)
among all features in Table~\ref{tab:single_feature_analysis}.  However,
as compared to those in Dataset A, most features have relatively low PCC
values in Dataset B.  To investigate further, we divide distorted images
into two groups. The first group includes additive noise and
transmission errors while the second group includes the Gaussian blur,
sampling blur, JPEG compression, and JPEG 2000 compression.  For the
first group, as the distortion level becomes higher, the MOS decreases
monotonically as shown in Fig.~\ref{fig:group1}. However, we cannot
observe such a coherent MOS movement trend for the second group in 
Fig.~\ref{fig:group2}. 

This phenomenon can be explained below. Differences among neighboring
pixels caused by the first group of distortions such as additive noise
or transmission errors are generally large, leading to scattered
geometric distortions in the rendered image especially around object
boundaries.  They tend to get worse as the distortion level increases.
On the other hand, differences among neighboring pixels caused by the
second group of distortions are often small and changing gradually. 
As a result, we may not be able to recognize them easily although geometric 
errors exist in the rendered image. \\

\noindent{\em Scorer \#9 - Formula-based Metric:}

Due to the weaker correlation between the MOS level and the distortion
level as shown in Fig. \ref{fig:group2}, it is diffucult to obtain high
prediction accuracy via a learning-based algorithm from features of the
input depth map directly.  To overcome this challenge, we exploit the
auxiliary information from two rendered views. That is, we use the SDP
index~\cite{cit:IVMSP} that is computed based on  
three depth maps as a candidate scorer. It helps boost the prediction accuracy as
presented in Section~\ref{sc:performance}. 

\subsection{ Learning-based Scorers and Fuser Design}\label{sec:4.3}

The proposed PBSIQA system composed by nine quality scorers is
summarized in Fig.~\ref{fig:Ensemble_model}.  To yield an intermediate
score from scorers \#1$\sim$\#8, we adopt the support vector regression
(SVR) technique.  Consider a set of training data $(\boldsymbol{x}_n,
y_n)$, where $\boldsymbol{x}_n$ is a feature vector and $y_n$ is the
target value, \textit{e.g.} the subjective quality score of the $n$th
image. In the $\varepsilon$-SVR \cite{cit:Smola2002}, \cite{cit:Basak2007}, the
objective is to find a mapping function $f(\boldsymbol{x}_n)$ that has a
deviation at most $\varepsilon$ from the target value, $y_n$, for all
training data. The mapping function is in form of
\begin{equation} \label{eq:SVR1}
f(\boldsymbol{x}) = \boldsymbol{w}^T \phi(\boldsymbol{x})+b,
\end{equation}
where $\boldsymbol{w}$ is a weighting vector, $\phi(\cdot)$ is a
non-linear function, and b is a bias term. We should find
$\boldsymbol{w}$ and $b$ satisfying the following condition:
\begin{equation} \label{eq:SVR2}
|f(\boldsymbol{x}_n)-y_n| \le \varepsilon, \quad \forall n=1,2,\ldots ,N_t,
\end{equation}
where $N_t$ is the number of training data.  

Although there exist several kernels such as linear, polynomial and sigmoid,
we use the radial basis function (RBF) in form of
\begin{equation} \label{eq:SVR6}
K(\boldsymbol{x_i},\boldsymbol{x_j}) = \exp(-\rho \|\boldsymbol{x_i}-
\boldsymbol{x_j}\|^2), \quad \rho > 0
\end{equation}
where $\rho$ is the radius controlling parameter, since it provides good
performance in applications \cite{cit:Chang2011}.  Furthermore, it is
not easy to determine a proper $\varepsilon$ value. Thus, we use a
different version of the regression algorithm called the $\nu$-SVR
\cite{cit:Basak2007}, where $\nu \in (0,1)$ is a control parameter to
adjust the number of support vectors and the accuracy level. In other
words, $\varepsilon$ becomes a variable in an optimization problem, and
we can obtain the same $f(\boldsymbol{x})$ and $\boldsymbol{w}$ more
conveniently. 

\begin{table*}[t]
\centering
\caption{Performance comparison among 6 sub-databases of 
Dataset C of the MCL3D database in terms of the
rank and PCC values (inside the parenthesis) with respect to each of the 
eight learning-based scorers, where the top-ranked sub-database is marked in
bold.}
    \begin{tabular}{c|>{\centering\arraybackslash}m{3cm}|cccccc}
    \toprule
    \textbf{Scorer \#} & \textbf{Target Distortion} & \textbf{\shortstack{AWN\\sub-database}} & \textbf{\shortstack{Gaussian Blur\\sub-database}} & \textbf{\shortstack{JP2K Coding\\sub-database}} 
     & \textbf{\shortstack{JPEG Coding\\sub-database}} & \textbf{\shortstack{Sampling Blur\\sub-database}} & \textbf{\shortstack{Trans. Loss\\sub-database}} \\ 
    \toprule
    \textbf{1} & \textbf{Blurring Distortion} & 3 (0.80) & \textbf{1 (0.89)} & 2 (0.84) & 4 (0.77) & 5 (0.70) & 6 (0.57) \\ \hline
    \textbf{2} & \textbf{Blocking Artifact} & 3 (0.81) & 6 (0.67) & 2 (0.82) & \textbf{1 (0.88)} & 4 (0.80) & 5 (0.70) \\ \hline
    \textbf{3} & \textbf{Additive Noise} & \textbf{1 (0.89)} & 5 (0.72) & 2 (0.77) & 4 0.75) & 6 (0.66) & 3 (0.76) \\ \hline
    \textbf{4} & \textbf{Global Structural Error} & \textbf{1 (0.90)} & 4 (0.78) & 3 (0.79) & 2 (0.84) & 6 (0.65) & 5 (0.66) \\ \hline
    \textbf{5} & \textbf{Local Structural Error} & \textbf{1 (0.91)} & 4 (0.79) & 3 (0.81) & 2 (0.82) & 5 (0.72) & 6 (0.66) \\ \hline
    \textbf{6} & \textbf{\shortstack{Luminance Change\\\& Scene Structural Error}} & 4 (0.73) & 5 (0.7) & 2 (0.84) & \textbf{1 (0.87)} & 3 (0.76) & 6 (0.66) \\ \hline
    \textbf{7} & \textbf{Transmission Error} & 2 (0.82) & 5 (0.64) & 6 (0.63) & 3 (0.68) & 3 (0.68) & \textbf{1 (0.84)} \\ \hline
    \textbf{8} & \textbf{Geometric Error} & \textbf{1 (0.92)} & 6 (0.39) & 4 (0.60) & 2 (0.70) & 5 (0.40) & 2 (0.70) \\
    \bottomrule
    \end{tabular}%
  \label{tab:ScorerAnalysis}%
\end{table*}%

\begin{table*}[t]
  \scriptsize
  \centering
  \caption{Cumulative performance improvement in terms of PCC via fusion against Dataset C of the MCL-3D database.}
    \begin{tabular}{c||c||cc|cc|cc|cc|cc|cc}
    \toprule
    \multirow{3}[0]{*}{\textbf{Sub-database}} & \multirow{3}[0]{*}{\textbf{\shortstack{Base\\Performance\\(\#8: Geometric\\Error)}}} & \multicolumn{2}{c|}{\multirow{2}[0]{*}{\textbf{\shortstack{Blurring\\Distortion\\(\#2)}}}} & \multicolumn{2}{c|}{\multirow{2}[0]{*}{\textbf{\shortstack{Additive\\
Noise\\(\#3)}}}} & \multicolumn{2}{c|}{\multirow{2}[0]{*}{\textbf{\shortstack{Blocking\\Artifact\\(\#1)}}}} & \multicolumn{2}{c|}{\multirow{2}[0]{*}{\textbf{\shortstack{Transmission\\Error\\(\#7)}}}} & \multicolumn{2}{c|}{\multirow{2}[0]{*}{\textbf{\shortstack{Structural\\Error\\(\#4 \#5)}}}} & \multicolumn{2}{c}{\multirow{2}[0]{*}{\textbf{\shortstack{Luminance\\Change\\(\#6)}}}} \\
          &       & \multicolumn{2}{c|}{} & \multicolumn{2}{c|}{} & \multicolumn{2}{c|}{} & \multicolumn{2}{c|}{} & \multicolumn{2}{c|}{} & \multicolumn{2}{c}{} \\\\
          &       & \textbf{} & \textbf{Perf.
Inc.} & \textbf{} & \textbf{Perf.
Inc.} & \textbf{} & \textbf{Perf.
Inc.} & \textbf{} & \textbf{Perf.
Inc.} & \textbf{} & \textbf{Perf.
Inc.} & \textbf{} & \textbf{Perf.
Inc.} \\
    \midrule
    \midrule

    AWN   & 0.92  & 0.78  & -17.9\% & 0.90  & \textbf{13.3\%} & 0.90  & 0.0\% & 0.90  & 0.0\% & 0.91  & 1.1\% & 0.92  & 1.1\% \\
    Gaussian Blur & 0.35  & 0.86  & \textbf{59.3\%} & 0.89  & 3.4\% & 0.93  & 4.3\% & 0.93  & 0.0\% & 0.94  & 1.1\% & 0.95  & 1.1\% \\
    JP2K  & 0.53  & 0.78  & 32.1\% & 0.87  & 10.3\% & 0.92  & 5.4\% & 0.92  & 0.0\% & 0.93  & 1.1\% & 0.95  & 2.1\% \\
    JPEG  & 0.81  & 0.79  & -2.5\% & 0.83  & 4.8\% & 0.93  & \textbf{10.8\%} & 0.94  & 1.1\% & 0.95  & 1.1\% & 0.97  & 2.1\% \\
    Sampling Blur & 0.45  & 0.84  & 46.4\% & 0.78  & -7.7\% & 0.87  & 10.3\% & 0.90  & 3.3\% & 0.91  & 1.1\% & 0.93  & 2.2\% \\
    Trans. Loss & 0.66  & 0.84  & 21.4\% & 0.79  & -6.3\% & 0.72  & -9.7\% & 0.83  & \textbf{13.3\%} & 0.87  & \textbf{4.6\%} & 0.91  & \textbf{4.4\%} \\
    \midrule
    Whole Dataset C & 0.58  & 0.74  & 21.6\% & 0.84  & 11.9\% & 0.89  & 5.6\% & 0.91  & 2.2\% & 0.92  & 1.1\% & 0.93  & 1.1\% \\
    \bottomrule
    \end{tabular}%
  \label{tab:progressive}%
\end{table*}%

At Stage II of the PBSIQA system, we fuse all intermediate scores from
the scorers at Stage I to determine the final quality score. We adopt
the $\nu$-SVR algorithm to implement the fuser. Suppose that there are
$n$ scorers with $m$ training stereoscopic image pairs. For the $i$th
training pair, we compute the intermediate quality score $s_{i,j}$,
where $i=1,2,...,m$ is the stereoscopic image pair index and
$j=1,2,...,9$ is the scorer index.  Let $\boldsymbol{s}_i=
(s_{i,1},s_{i,2},\cdots,s_{i,9})$ be the intermediate score vector for
the $i$th image pair. We train the fuser using $\boldsymbol{s_i}$ with
all image pairs in the training set and determine the weighting vector
$\boldsymbol{w}$ and the bias parameter $b$ accordingly. Finally, the
PBSIQA-predicted quality metric for a given stereoscopic image pair in
the test set can be found via
\begin{equation} \label{eq:ProposedIndex}
Q(\boldsymbol{s}) = \boldsymbol{w}^T \phi(\boldsymbol{s})+b.
\end{equation}

\subsection{Training and Test Procedures}\label{sc:learning_procedure}

The $n$-fold cross validation~\cite{cit:CV} is a common strategy to
evaluate the performance of a learning-based algorithm to ensure
reliable results and prevent over-fitting, where the data are split into
$n$ chunks, and one chunk is used as the test data while the remaining
$n-1$ chunks are used as training data.  
We ensured that the model is dominantly trained by the
different distortion types from that of testing data since it may boost the 
prediction accuracy.
The same experiment is repeated
$n$ times by employing each of the $n$ chunks as the test data.
Finally, the overall performance is determined 
over all the predicted scores. In the proposed framework, training
data are used to generate a regression model of each scorer at Stage I.
Then, a regression model of the fuser is obtained by training all
intermediate scores from Stage I as input features with $n$-fold cross
validation at Stage II, where the number of samples is the same as the
number of stereoscopic image pairs in the $n-1$ chunks. In all reported
experiment results, we use the 10-fold cross validation.  In addition, a
feature scaling operation is performed before the training and test
processes.  It is conducted to avoid features of a larger numeric range
dominating those of a smaller numeric range. For example, the PSNR value
has a larger range of values than the other two features in scorer \#3.
We scale the feature values of each scorer to the unit range of [0,1] at
Stage I. 

At the training stage, our goal is to determine the optimal weighting
vector $\boldsymbol{w}$ and bias $b$ that minimize the error between 
MOS and the predicted score, i.e.,
\begin{equation} \label{eq:Training}
\sum_i |\mbox{MOS}_i-Q(\boldsymbol{s}_i)|^2.
\end{equation}
Since we use RBF as a kernel function, two parameters $C$ and $\gamma$
should be optimized to achieve the best regression accuracy. We conduct
parameter search on $C$ and $\gamma$ at the training stage using the
cross validation scheme in \cite{cit:Chang2011}. Various pairs of $(C,
\gamma)$ are tried, and the one with the best cross validation accuracy
is selected.  After the best $(C, \gamma)$ pair is found, the whole
training set is used again to generate the final scorer. 

At the test stage, we use the intermediate score vector $s_i$ in Eq.
(\ref{eq:ProposedIndex}) to determine the quality score. The test can be
quickly done since all model parameters are decided at the training
stage. 

\section{Performance Evaluation}\label{sc:performance}

To evaluate the performance of the proposed PBSIQA metric, 
we follow the suggestions of ITU-T(P.1401)~\cite{cit:ITU1401} and use three
performance measures: 1) the Pearson correlation coefficient (PCC) to measure 
the linear relationship between a model's performance and the subjective data , 
2) the Spearman rank-order correlation coefficient (SROCC) for the prediction 
monotonicity, and 3) the root mean squared error (RMSE) for the prediction 
accuracy. Before calculating the performance measures, we apply the
monotonic logistic function to the predicted quality scores so as to fit
the subjective quality scores (MOS) and remove any nonlinearity via
\begin{equation} \label{eq:Logistic}
f(s) = \frac{\beta_1-\beta_2}{1+exp(-\frac{s-\beta_3}{|\beta_4|})}+\beta_2,
\end{equation}
where $s$ and $f(s)$ are the predicted score and the fit predicted
score, respectively, and $\beta_k$ ($k=1,2,3,4$) are the parameters to
minimize the mean squared error between $f(s)$ and MOS. In addition, 
this mapping function compensates for offsets, different biases, and 
other shifts between the scores, without changing the rank order.

\subsection{Performance Comparison of Stage I Scorers}

First, we compare the performance of eight learning-based scorers
(Scorers \#1-8) and show that each scorer truly offer good performance
for its target distortion type. Here, we consider Datset
C of the MLC-3D database only, and divide it into six sub-databases so
that each contains one distortion type as described in
Section~\ref{sc:database}. The six distortion types are listed in the
top row of Table~\ref{tab:ScorerAnalysis}.  We compute the PCC value of
each scorer against each sub-database and rank its effectiveness in the
descending order of PCC values individually. Table
\ref{tab:ScorerAnalysis} summarizes the results of scorers in Stage I
only (namely, without fusion in Stage II). 
The performance of scorers \#1, \#2, and \#3 matches well with their
target design. First, scorer \#1 has the highest correlation on the
blurred and JPEG2000 compression sub-databases. This is reasonable
since, besides the ringing artifect, the main distortion of the JPEG2000
compression sub-database is blurring. Scorer \#2 is designed for the
blocking artifact, and the JPEG compression database has the top rank.
Scorer \#3 has the best performance on additive noise database as
designed. 

Scorer \#4 (for global structural error) and scorer \#5 (for local
structural error) share strong similarity in their rank orders.
Specifically, they provide the first and second best performance in
additive noise and JPEG compressed databases. As observed in
\cite{cit:IVMSP}, the information additive distortion (IAD) is more
obvious than the information loss distortion (ILD) among all structural
distortions.  Additive noise and blockiness are typical exmaples of IAD.
Our observation is consistent with that in \cite{cit:IVMSP}.  Scorer \#6
yields the best result for the JPEG compressed sub-database since we may
see color/luminance chages if the compression ratio is too high, which
can be captured by the singular value feature of Scorer \#6.  However,
it has poor correlation with human subjective experience on the additive
noise sub-database. This implies that, although scorers \#4, \#5 and \#6
are designed for structural errors in common, they capture different
aspects. 

Scorer \#7 offers the best result for transmission errors to meet the
target design. Last, scorer \#8, designed to capture geometric
distortions, provides the best result in databases with additve noise
and transmission errors, which was already explained in Section
\ref{sc:scorerdesign_depth}.

\begin{table*}[htbp]
  \centering
  \caption{Performance comparison over the MCL-3D database.}
    \begin{tabular}{rr|ccc|ccc|ccc}
    \toprule
          &       & \multicolumn{3}{c|}{\textbf{Dataset A}} & \multicolumn{3}{c|}{\textbf{Dataset B}} & \multicolumn{3}{c}{\textbf{Dataset C}} \\ [1ex]
    \multicolumn{2}{c|}{\textbf{Quality Indices}} & \textbf{PCC} & \textbf{SROCC} & \textbf{RMSE} & \textbf{PCC} & \textbf{SROCC} & \textbf{RMSE} & \textbf{PCC} & \textbf{SROCC} & \textbf{RMSE} \\ [0.9ex] \hline \hline
    \multicolumn{2}{c|}{\textbf{Proposed PBSIQA}} & \textbf{0.921 } & \textbf{0.913 } & \textbf{0.703 } & \textbf{0.848 } & \textbf{0.775 } & \textbf{0.580 } & \textbf{0.933 } & \textbf{0.920 } & \textbf{0.741 } \\ [1ex] \hline
    \multicolumn{1}{c|}{\multirow{5}[4]{*}{\textbf{3D indices}}} 
                                    & \multicolumn{1}{c|}{\textbf{BQPNR}} & 0.591 & 0.533 & 2.241 & 0.675  & 0.512  & 1.181  & 0.617 & 0.522 & 2.339  \\[0.9ex]
    \multicolumn{1}{c|}{} & \multicolumn{1}{c|}{\textbf{CYCLOP}} & 0.742  & 0.730  & 1.758  & 0.805  & 0.730  & 0.981  & 0.791  & 0.772  & 1.513  \\ [0.9ex]
    \multicolumn{1}{c|}{} & \multicolumn{1}{c|}{\textbf{RKS}} & 0.710  & 0.679  & 1.578  & \textbf{0.829}  & \textbf{0.748}  & \textbf{0.894}  & 0.731  & 0.731  & 1.594  \\[0.9ex]
    \multicolumn{1}{c|}{} & \multicolumn{1}{c|}{\textbf{Benoit}} & 0.560  & 0.559  & 1.855  & 0.698  & 0.533  & 1.147  & 0.593  & 0.593  & 1.884  \\[0.9ex]
    \multicolumn{1}{c|}{} & \multicolumn{1}{c|}{\textbf{Campisi}} & 0.752  & 0.755  & 1.422  & 0.815  & 0.676  & 0.926  & 0.824  & 0.824  & 1.323  \\ [0.9ex]\hline
    \multicolumn{1}{c|}{\multirow{12}[6]{*}{\textbf{2D indices}}} & \multicolumn{1}{c|}{\textbf{C4}} & \textbf{0.788 } & \textbf{0.796 } & \textbf{1.402 } & 0.818  & 0.676  & 0.921  & 0.824  & 0.833  & 1.323  \\[0.9ex]
    \multicolumn{1}{c|}{} & \multicolumn{1}{c|}{\textbf{IFC}} & 0.639  & 0.593  & 1.722  & 0.662  & 0.160  & 0.199  & 0.671  & 0.609  & 1.733  \\[0.9ex]
    \multicolumn{1}{c|}{} & \multicolumn{1}{c|}{\textbf{MSE}} & 0.673  & 0.675  & 1.657  & 0.785  & 0.739  & 0.991  & 0.709  & 0.703  & 1.648  \\[0.9ex]
    \multicolumn{1}{c|}{} & \multicolumn{1}{c|}{\textbf{NQM}} & 0.754  & 0.787  & 1.470  & 0.766  & 0.672  & 1.030  & \textbf{0.836 } & \textbf{0.836 } & \textbf{1.282 } \\[0.9ex]
    \multicolumn{1}{c|}{} & \multicolumn{1}{c|}{\textbf{PSNR\_HVS}} & 0.749  & 0.795  & 1.484  & 0.811  & 0.753  & 0.936  & 0.815  & 0.807  & 1.354  \\[0.9ex]
    \multicolumn{1}{c|}{} & \multicolumn{1}{c|}{\textbf{PSNR}} & 0.720  & 0.678  & 1.555  & 0.798  & 0.737  & 0.965  & 0.725  & 0.705  & 1.609  \\[0.9ex]
    \multicolumn{1}{c|}{} & \multicolumn{1}{c|}{\textbf{SNR}} & 0.705  & 0.699  & 1.589  & 0.806  & 0.738  & 0.947  & 0.743  & 0.721  & 1.566  \\[0.9ex]
    \multicolumn{1}{c|}{} & \multicolumn{1}{c|}{\textbf{SSIM}} & 0.533  & 0.534  & 1.896  & 0.785  & 0.728  & 0.992  & 0.575  & 0.587  & 1.914  \\[0.9ex]
    \multicolumn{1}{c|}{} & \multicolumn{1}{c|}{\textbf{UQI}} & 0.475  & 0.502  & 1.971  & 0.769  & 0.659  & 1.022  & 0.548  & 0.545  & 1.956  \\[0.9ex]
    \multicolumn{1}{c|}{} & \multicolumn{1}{c|}{\textbf{VIF}} & 0.663  & 0.656  & 1.677  & 0.746  & 0.582  & 1.066  & 0.696  & 0.671  & 1.678  \\[0.9ex]
    \multicolumn{1}{c|}{} & \multicolumn{1}{c|}{\textbf{VIFP}} & 0.604  & 0.622  & 1.784  & 0.743  & 0.602  & 1.072  & 0.665  & 0.646  & 1.744  \\[0.9ex]
    \multicolumn{1}{c|}{} & \multicolumn{1}{c|}{\textbf{VSNR}} & 0.693  & 0.720  & 1.615  & 0.790  & 0.730  & 0.981  & 0.760  & 0.756  & 1.519  \\[0.9ex]
    \bottomrule
    \end{tabular}%
  \label{tab:perf_comparison}%
\end{table*}%

\subsection{Performance Improvement via Fusion}
\begin{figure}[t]
\centering
\includegraphics[width=0.3\textwidth]{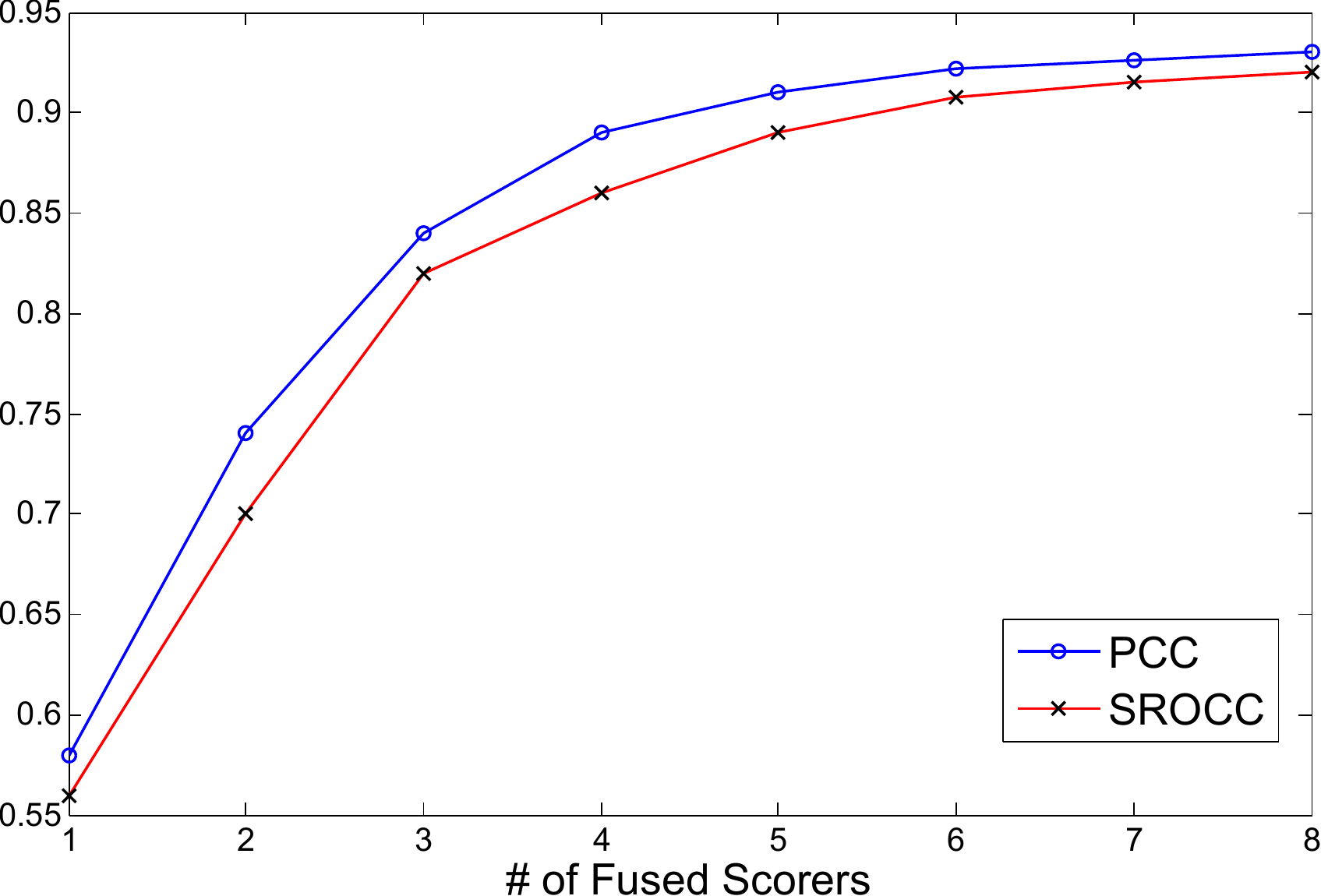}
\caption{The PCC and SROCC performance curves as a function of the
number of fused scorers in the PBSIQA system.}\label{fig:progressive}
\end{figure}

We show how the PBSIQA system can improve QA performance by fusing the
results from scorers at Stage I progressively in this section.  To
illustrate the design methodology, without loss of generality, we use
the performance of scorer \#8 for the geometric error as the base,
add one scorer at a time to account for the most
challenging distortion type in the remaining set, and show how the
added scorer improves the performance for six sub-databases and the
entire MCL-3D database. The results conducted against Dataset C of the
MCL-3D database are shown in Table \ref{tab:progressive}, where the base
performance with scorer \#8 is shown in the first column and the results
from scorers \#2, \#3, \#1, \#7, \#4, \#5, and \#6 are fused to its
score one by one cumulatively.  We see a substantial performance gain
for each sub-database when its associated scorer is included in the
fusion process. The whole database contains different distortion types
so that it is beneficial to fuse all scorers to get robust and accurate
predicted quality scores as shown in the last row of Table \ref{tab:progressive}. 
The design methodology (namely, the fusion order of scorers \#2, \#3,
\#1, \#7, \#4 and \#5 and, finally, \#6) is explained below.  The
performance gain for a given database is calculated in percentages as
\begin{equation} \label{eq:PerformanceGain}
\mbox{Performance\_Increase} = \frac{V_{\rm curr}-V_{\rm prev}}{V_{\rm prev}}
\times 100,
\end{equation}
where $V_{\rm curr}$ is the value of PCC after adding a
new scorer, while $V_{\rm prev}$ is the result of the previous stage.
Since the error rate of base scorer \#8 is highest in sub-databases with
Gaussian/sampling blur, we fuse scorer \#2 to scorer \#8 to reduce
erroneous prediction caused by blurring distortion. We show the
performance improvement for each sub-database in the second column of
Table \ref{tab:progressive} and see about 60\% and 45\% performance
gains for sub-databases with the Gaussian and the sampling blur,
respectively.  Although the performance gain drops in some
sub-databases, they are not as significant. As a result, we achieve a
performance gain of 21.6\% for the entire database. 
After the fusion of scorers \#8 and \#2, it is shown in Table
\ref{tab:progressive} that the system does not perform well against
additive noise. Thus, we fuse scorer \#3 to the system in the next step.
After the fusion of scorers \#8, \#2 and \#3, we observe a significant
performance boost (13\%) for the additive noise sub-database, and a
performance gain of 11.9\% for the entire database.  By following the
same line of thought, we can fuse more scorers and obtain better
performance for the entire database.  Based on the above discussion,
both PCC and SROCC values with respect to the whole MCL-3D database are
plotted as a function of the number of fused scorers in Fig.~\ref{fig:progressive}. 
The performance of the proposed PBSIQA system may
reach a saturation point if the number of fused scorers is sufficiently
larger.  It should also be emphasized that the proposed PBSIQA system
can be extended systematically. That is, if new distortion types are
introduced, we can design scorers to tailor to them and fuse new scorers
into the system.  In contrast, the traditional formula-based quality
metric does not have such flexibility. 

\subsection{Performance Comparison with Other Quality Indices}
\label{sc:Performance Comparison}

In this section, we compare the performance of the proposed PBSIQA metric
with other QA metrics against Datasets A, B, and C of MCL-3D, IVC-A and
LIVE-A.  The latter two are more challenging than their sources,
IVC~\cite{cit:IVC} and LIVE~\cite{cit:LIVE}, by including asymmetric
distortions.  Futhermore, we conduct a cross-database learning procedure
to demonstrate the robustness of the proposed PBSIQA metric.  That is,
the PBSIQA trained with the MCL-3D database is used to predict the
quality of stereo image pairs in IVC-A and LIVE-A. 
For performance benchmarking, we consider both 3D and 2D quality indices
in the literature.  The 3D indices include those denoted by
Benoit~\cite{cit:Benoit2008}, Campisi~\cite{cit:Campisi2007}, 
RKS~\cite{cit:RKS2012} and BQPNR~\cite{cit:Ryu2014}. The 2D
indices include: the Signal to Noise Ratio (SNR), the Peak Signal to
Noise Ratio (PSNR), the Mean Square Error (MSE), the Noise Quality
Measure~\cite{cit:NQM} (NQM), the Universal Quality
Index~\cite{cit:Wang2002} (UQI), the Structural Similarity
Index~\cite{cit:SSIM} (SSIM), the pixel-based VIF~\cite{cit:VIFP}
(VIFP), the visual signal-to-noise ratio~\cite{cit:VSNR} (VSNR), the
Peak Signal to Noise Ratio taking into account CSF~\cite{cit:PSNR-HVS}
(PSNR-HVS), C4~\cite{cit:C4}, and the image fidelity
criterion~\cite{cit:IFC} (IFC).  Since 2D metrics are only applied to a
single image, we use the average score of left and right rendered views
to yield the overall quality metric. 
We first focus on performance comparison against the MCL-3D database.
The results are summarized in Table~\ref{tab:perf_comparison}, where the
best and second best metrics in each column of are shown in bold. We
have the following observations.
\begin{itemize}
\item Dataset A of MCL-3D (texture distortion only) \\
The PBSIQA system has the best performance in terms of PCC, SROCC and
RMSE.  It outperforms all other 2D and 3D indices by a significant
margin.  Among the 2D indices, those designed by considering HVS offer
better performance such as C4, NQM, and PSNR\_HVS. 
\item Dataset B of MCL-3D (depth map distortion only) \\
The PBSIQA system gives the best performance in PCC, SROCC and RMSE, yet
its performance gain over the second best one narrows. Note that there
are only two scorers (scorers \#8 and \#9) in the PBSIQA system dedicated
to the depth map distortion. 
\item Dataset C of MCL-3D (both texture and depth map distortions) \\
The PCC and the SROCC values of the PBSIQA metric are 0.933 and 0.920,
respectively. It outperforms all benchmarking indices by a significant
margin. 
\end{itemize}

\begin{figure*}
        \centering
        \begin{subfigure}[t]{0.175\textwidth}
                \centering
                \includegraphics[width=\textwidth]{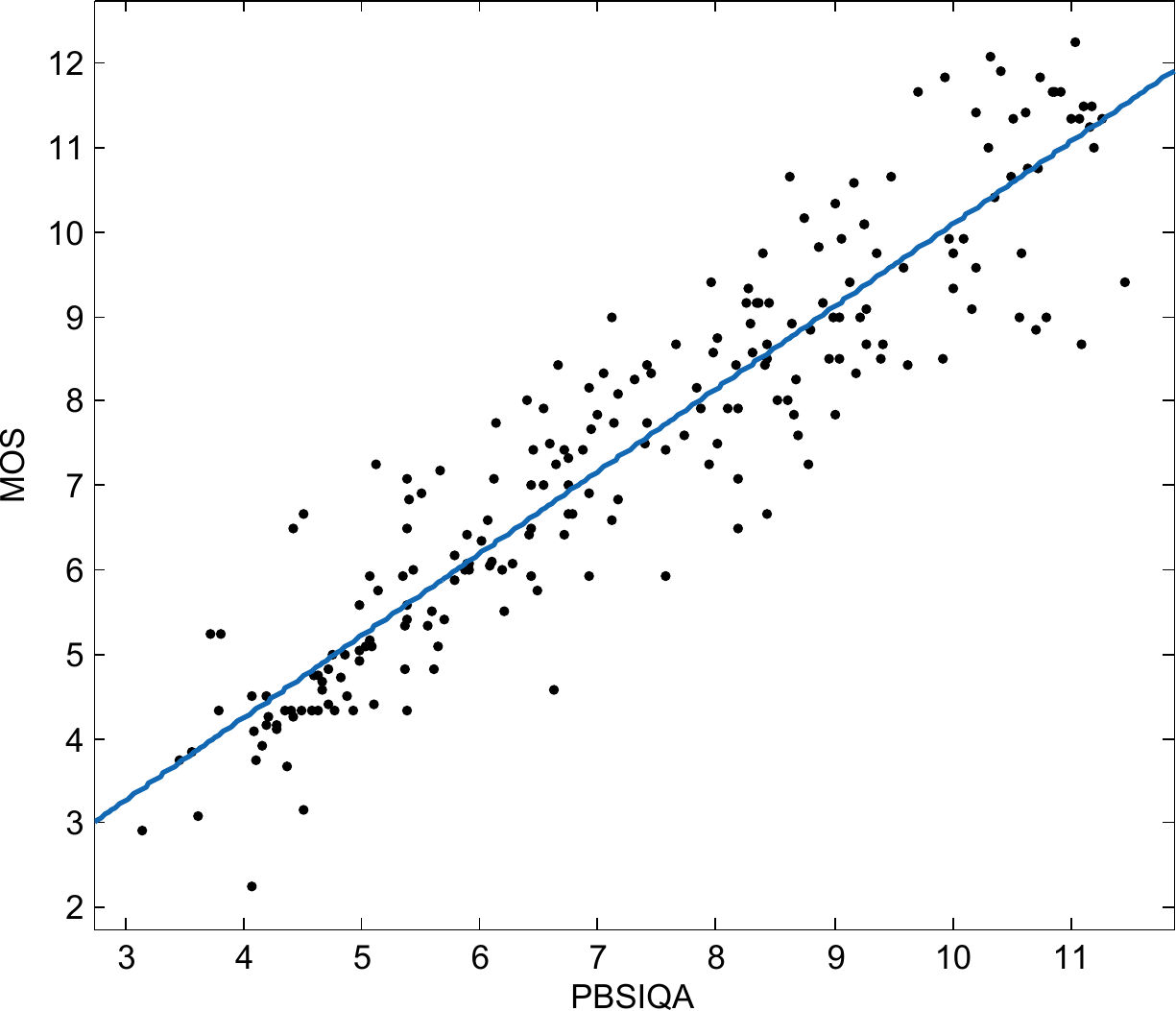}
                \caption{PBSIQA}
        \end{subfigure}
		\quad
        \begin{subfigure}[t]{0.175\textwidth}
                \centering
                \includegraphics[width=\textwidth]{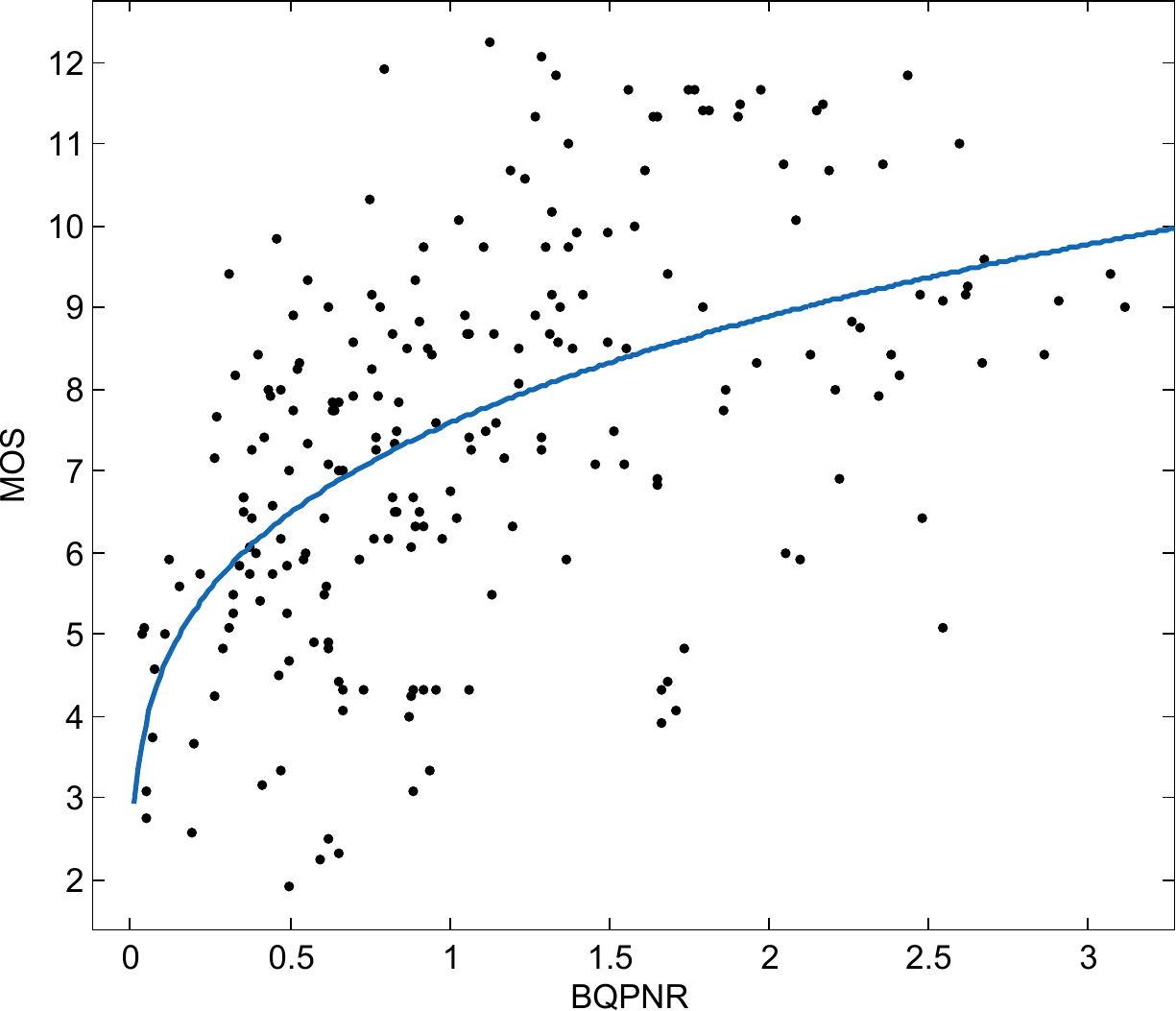}
                \caption{BQPNR}
        \end{subfigure}
		\quad
        \begin{subfigure}[t]{0.175\textwidth}
                \centering
                \includegraphics[width=\textwidth]{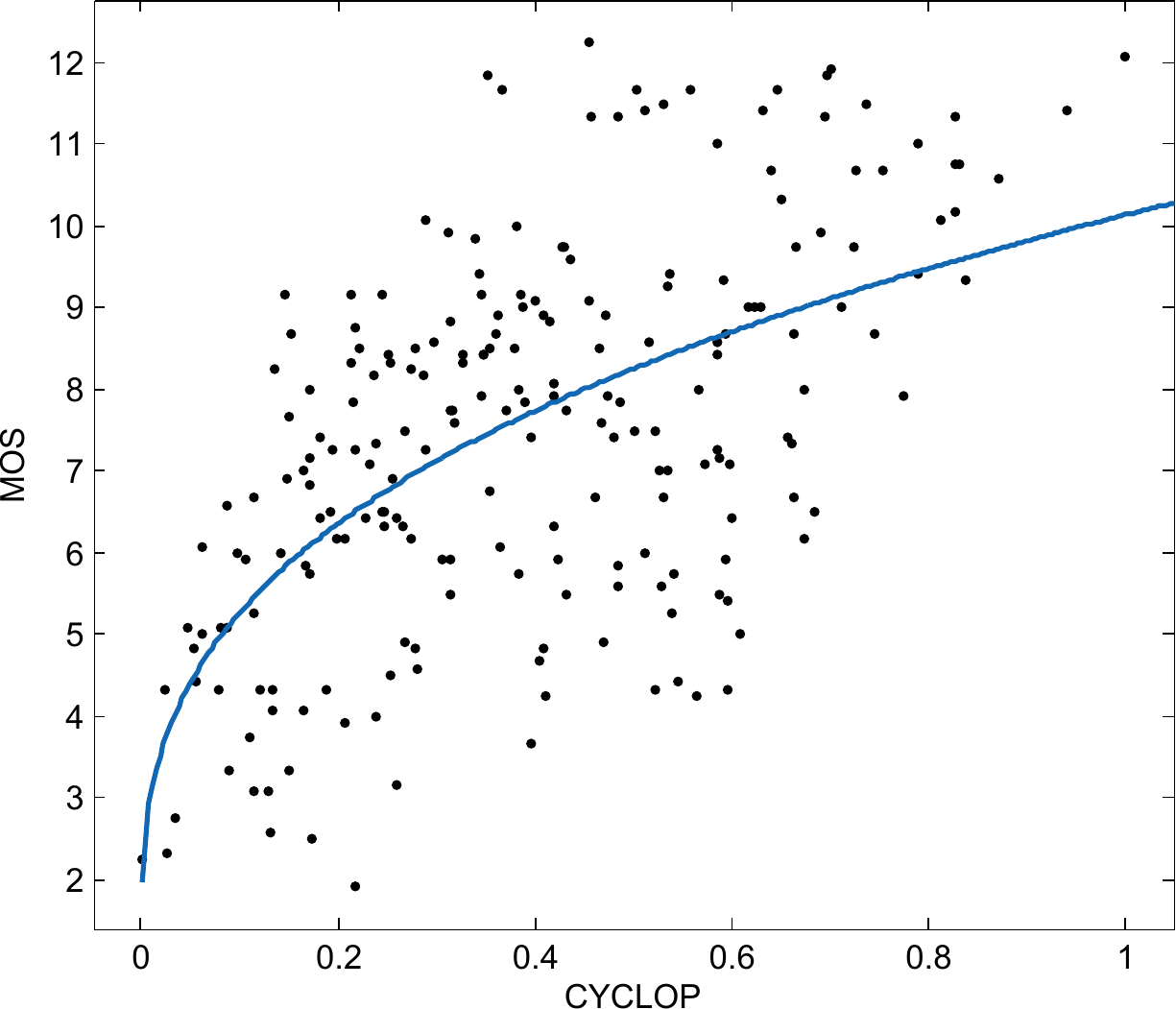}
                \caption{CYCLOP}
        \end{subfigure}
		\quad
        \begin{subfigure}[t]{0.175\textwidth}
                \centering
                \includegraphics[width=\textwidth]{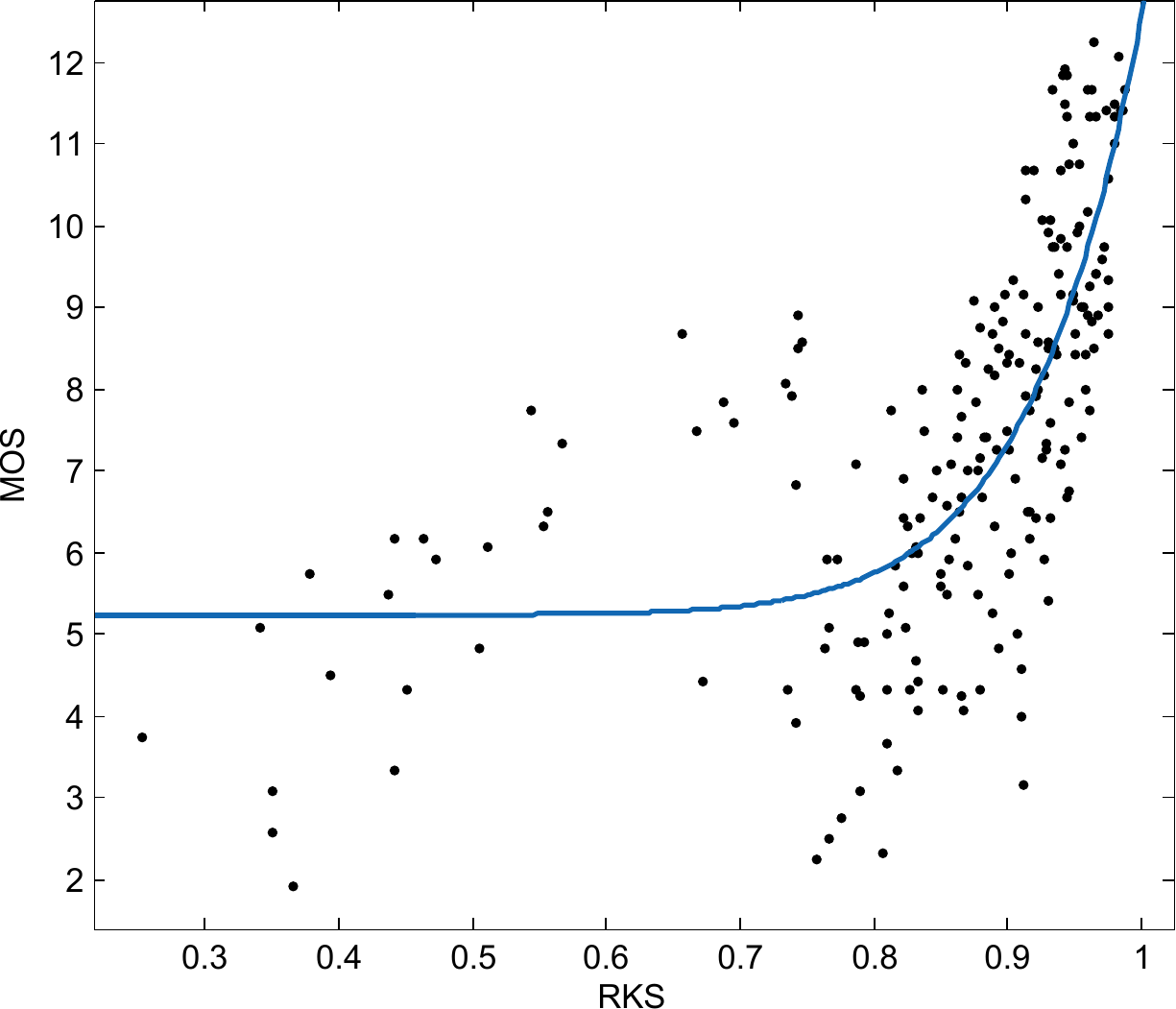}
                \caption{RKS}
        \end{subfigure}
		\quad
        \begin{subfigure}[t]{0.175\textwidth}
                \centering
                \includegraphics[width=\textwidth]{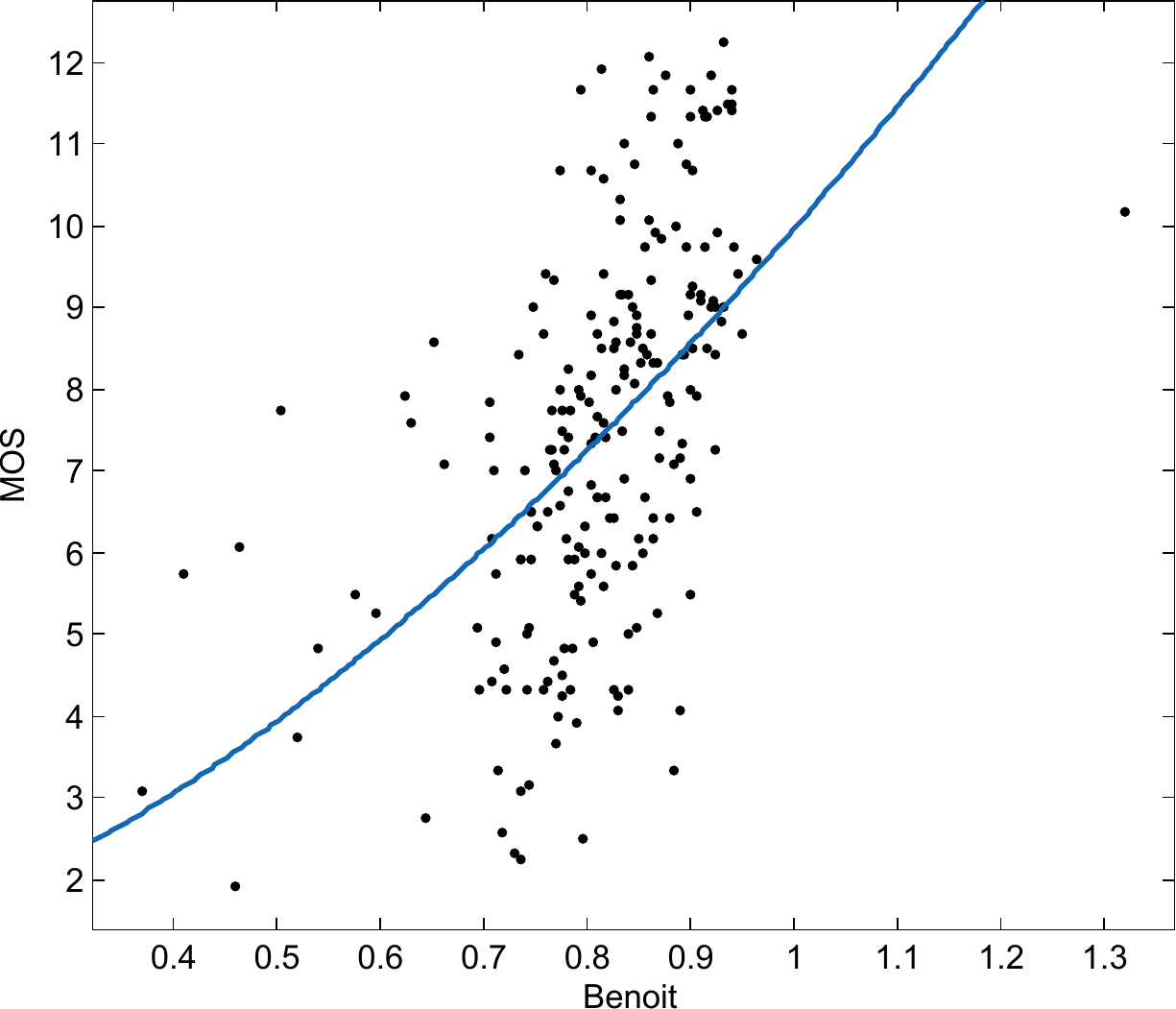}
                \caption{Benoit}
        \end{subfigure}
		\quad
        \begin{subfigure}[t]{0.175\textwidth}
                \centering
                \includegraphics[width=\textwidth]{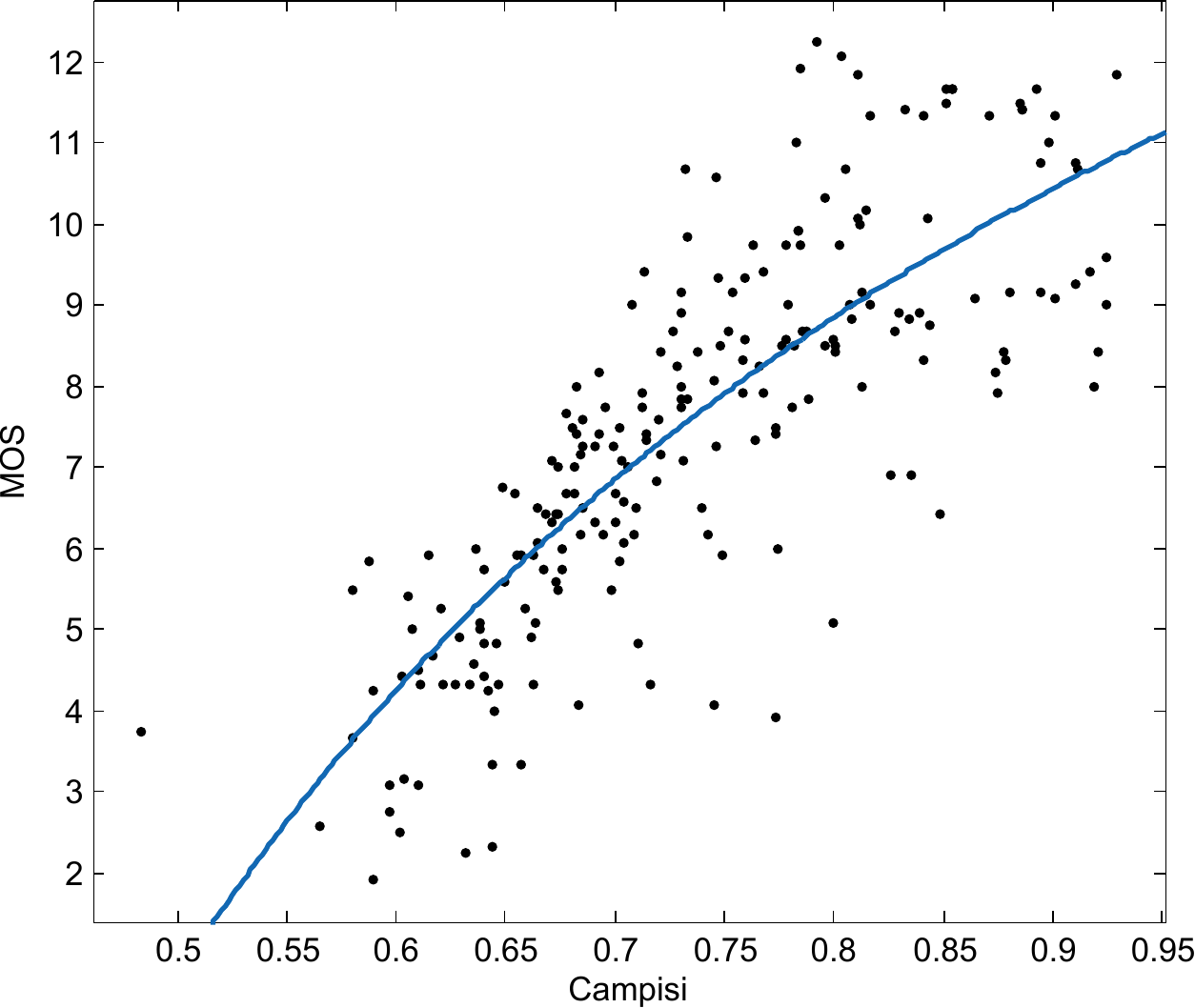}
                \caption{Campisi}
        \end{subfigure}
		\quad
        \begin{subfigure}[t]{0.175\textwidth}
                \centering
                \includegraphics[width=\textwidth]{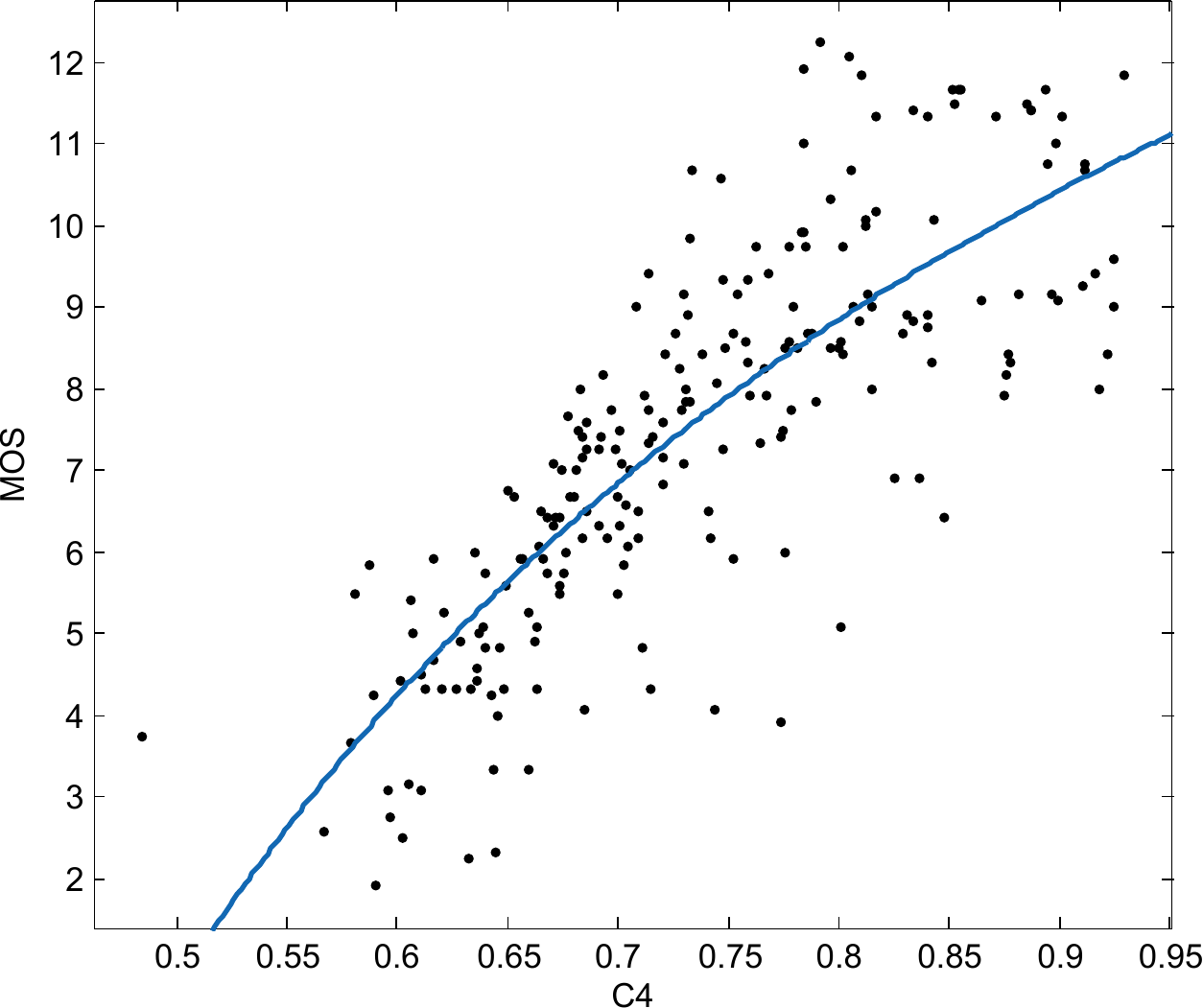}
                \caption{C4}
        \end{subfigure}
		\quad
        \begin{subfigure}[t]{0.175\textwidth}
                \centering
                \includegraphics[width=\textwidth]{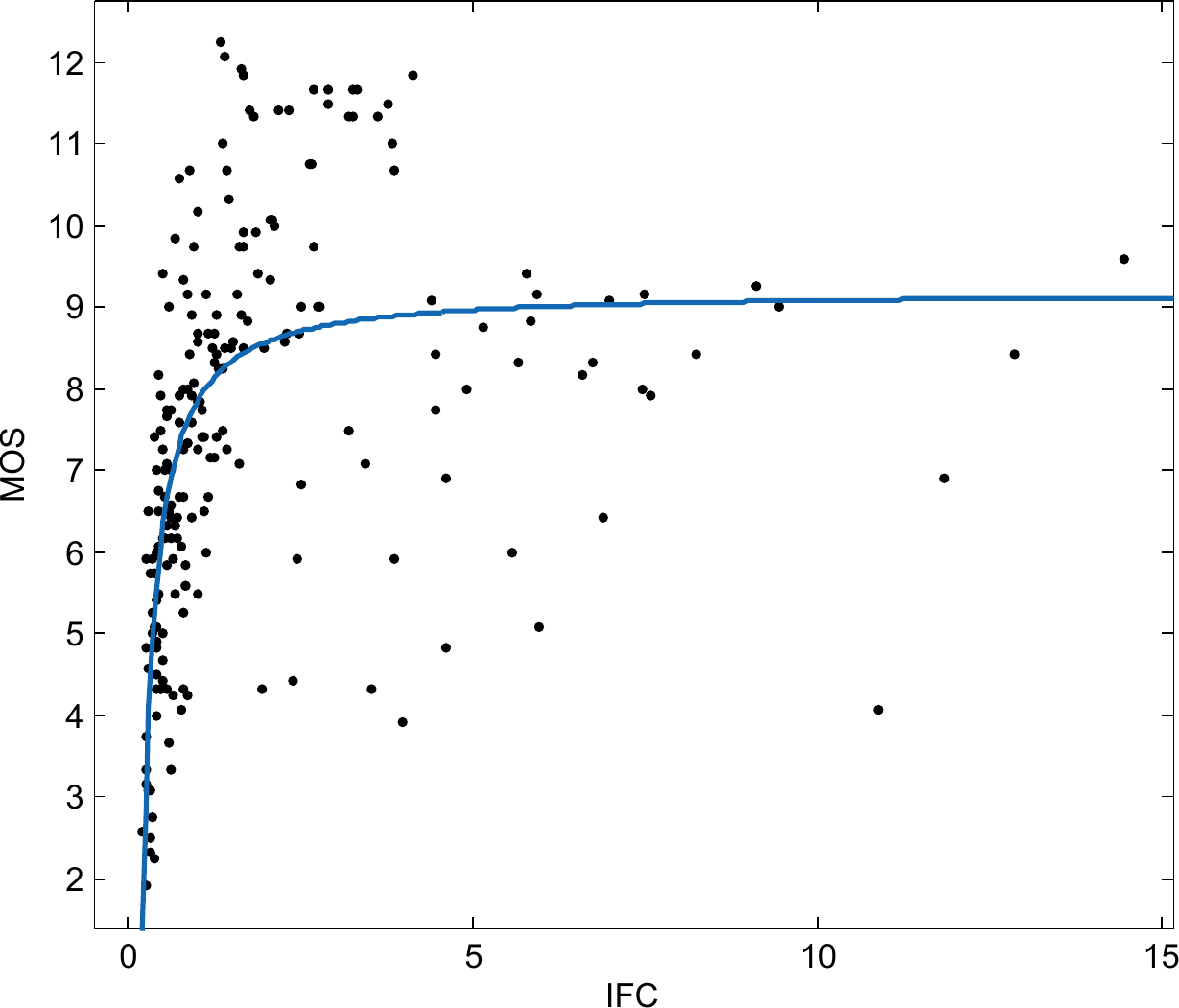}
                \caption{IFC}
        \end{subfigure}
		\quad
        \begin{subfigure}[t]{0.175\textwidth}
                \centering
                \includegraphics[width=\textwidth]{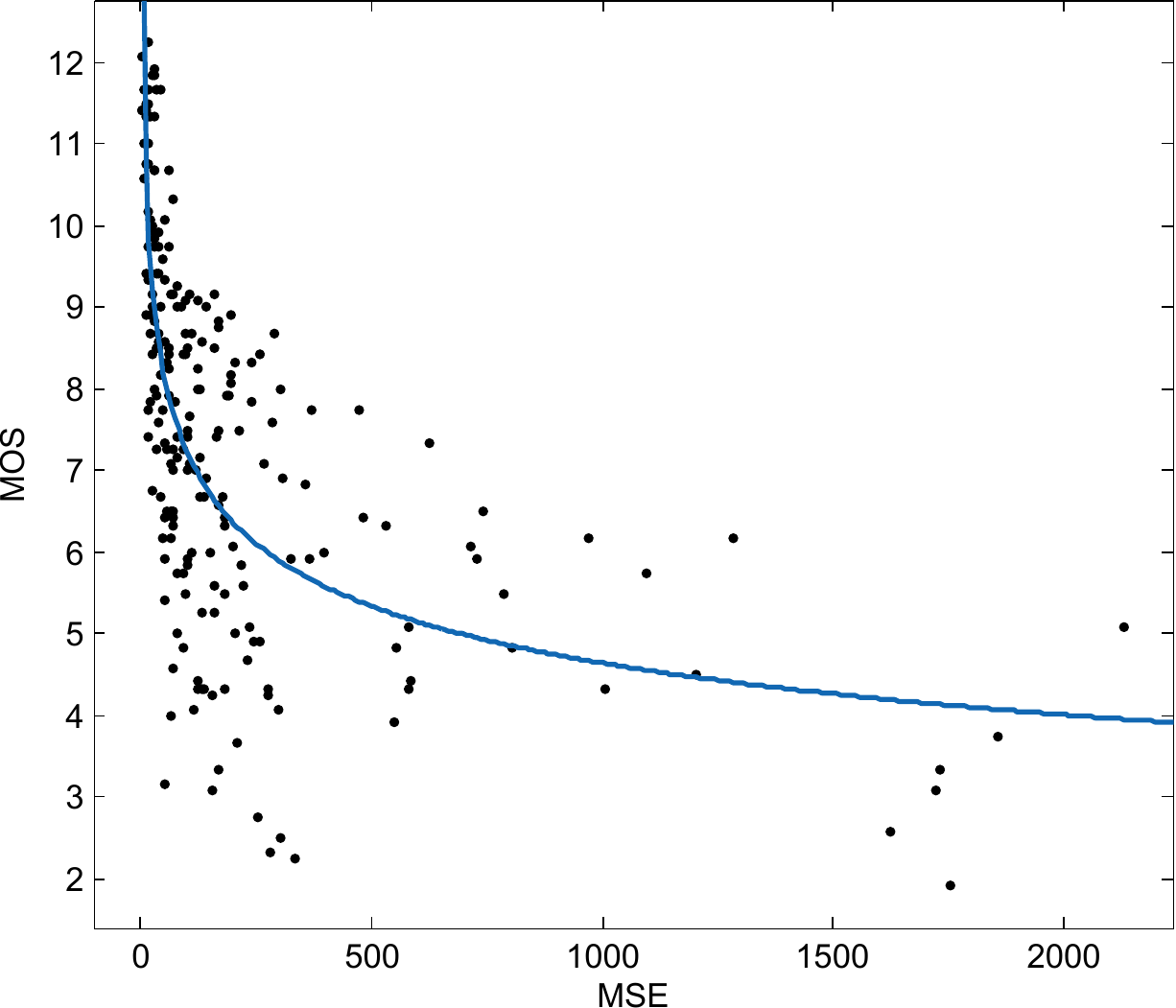}
                \caption{MSE}
        \end{subfigure}
		\quad
        \begin{subfigure}[t]{0.175\textwidth}
                \centering
                \includegraphics[width=\textwidth]{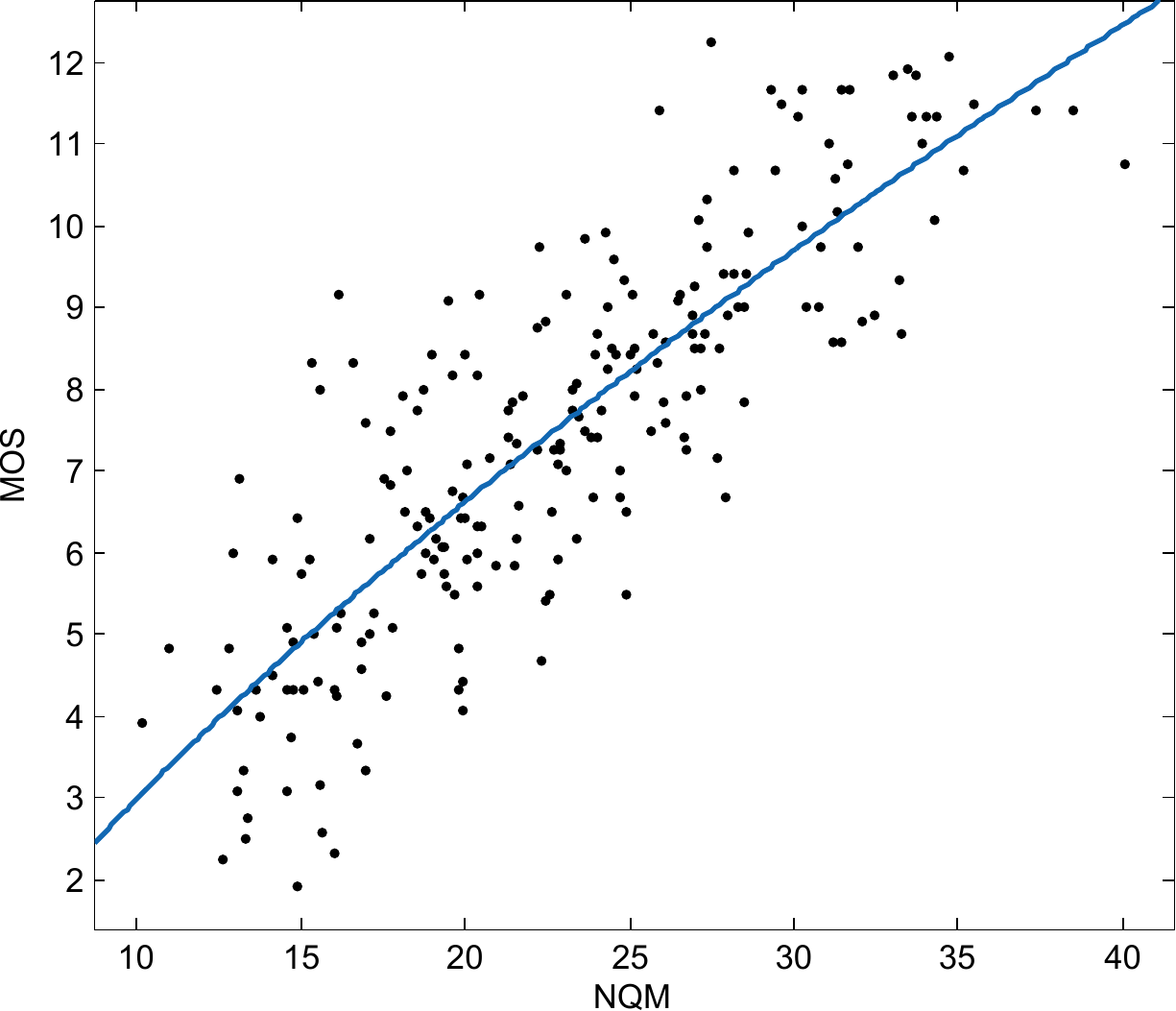}
                \caption{NQM}
        \end{subfigure}
		\quad
        \begin{subfigure}[t]{0.175\textwidth}
                \centering
                \includegraphics[width=\textwidth]{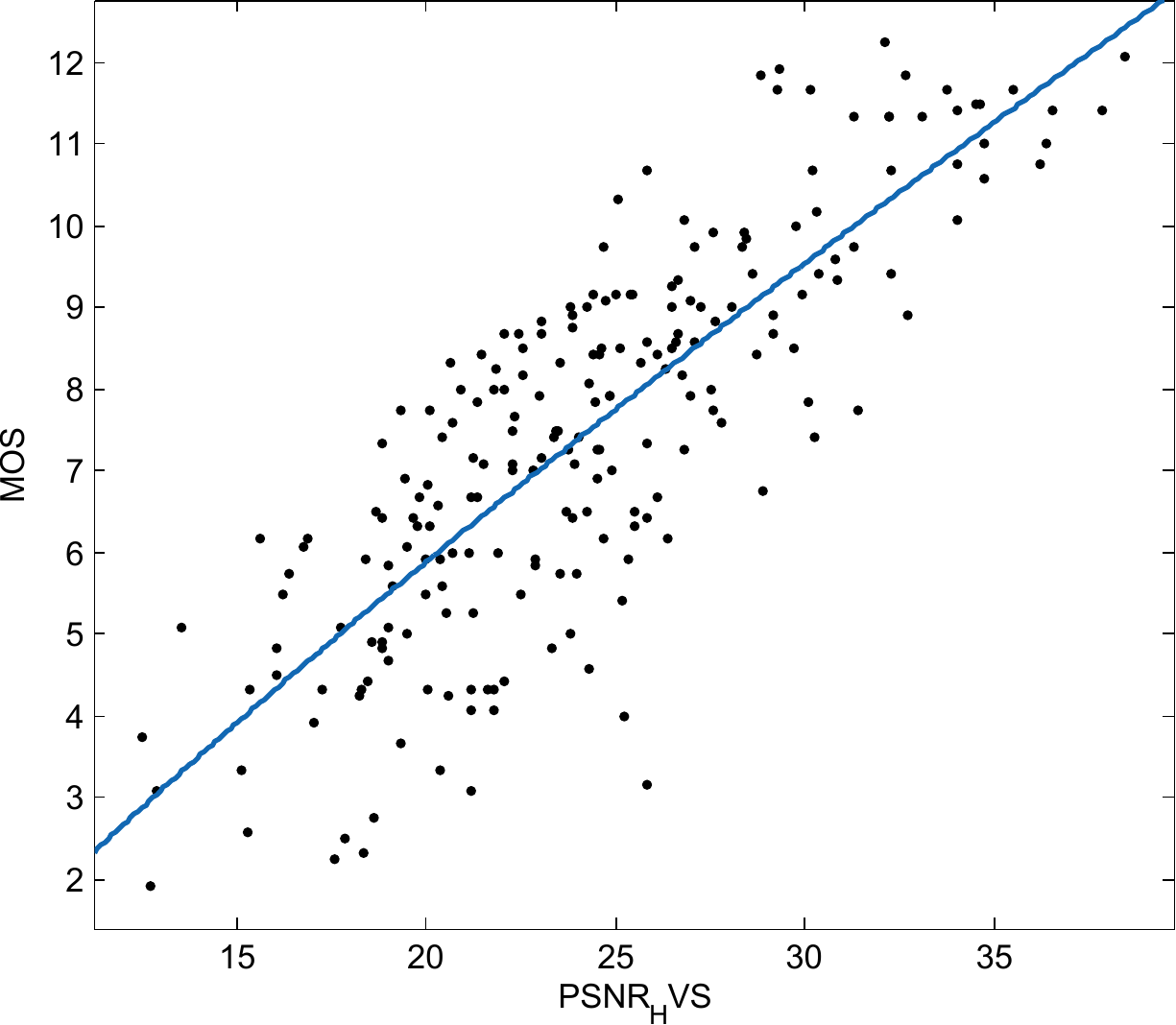}
                \caption{PSNR\_HVS}
        \end{subfigure}
		\quad
        \begin{subfigure}[t]{0.175\textwidth}
                \centering
                \includegraphics[width=\textwidth]{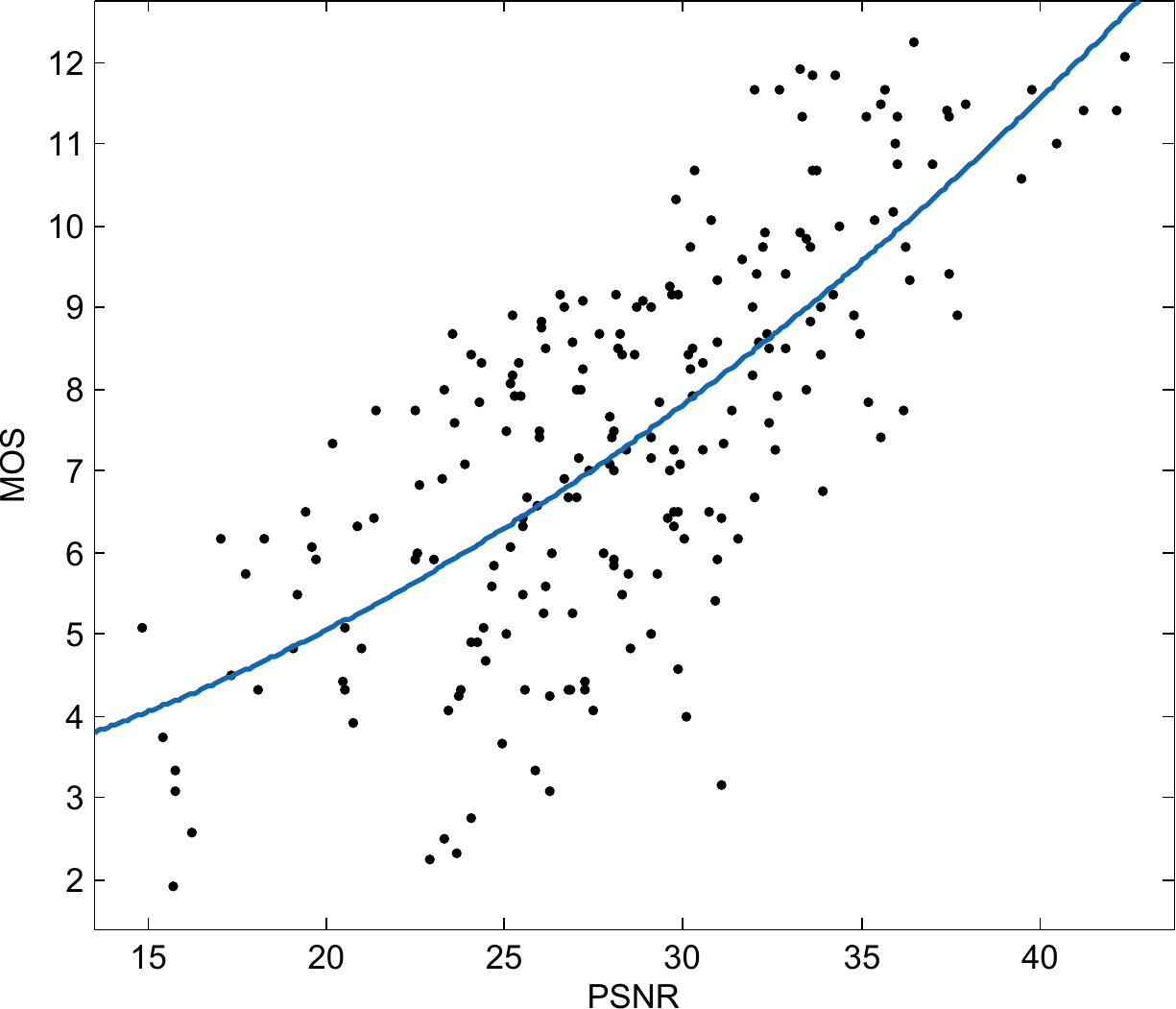}
                \caption{PSNR}
        \end{subfigure}
		\quad
        \begin{subfigure}[t]{0.175\textwidth}
                \centering
                \includegraphics[width=\textwidth]{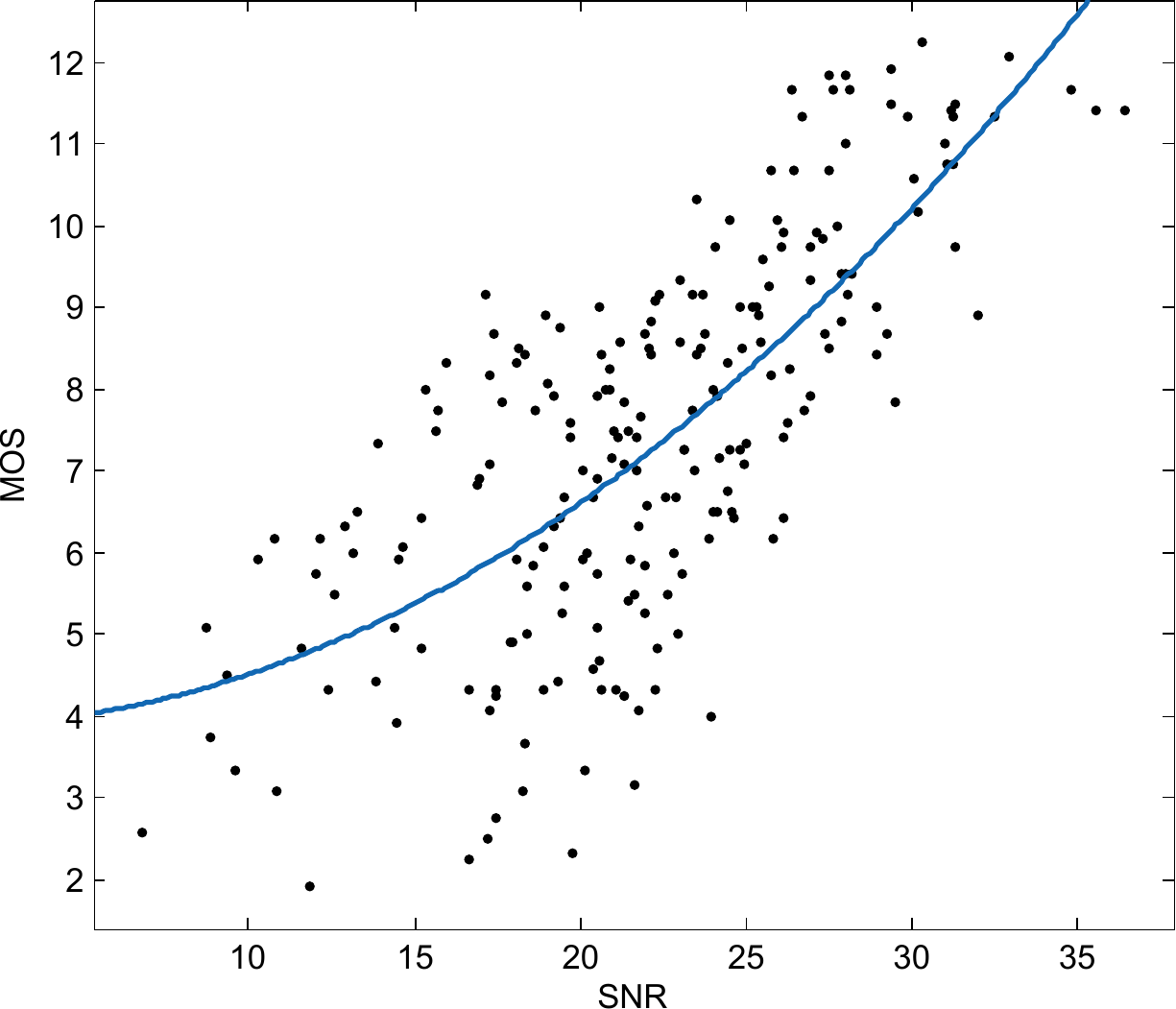}
                \caption{SNR}
        \end{subfigure}
		\quad
        \begin{subfigure}[t]{0.175\textwidth}
                \centering
                \includegraphics[width=\textwidth]{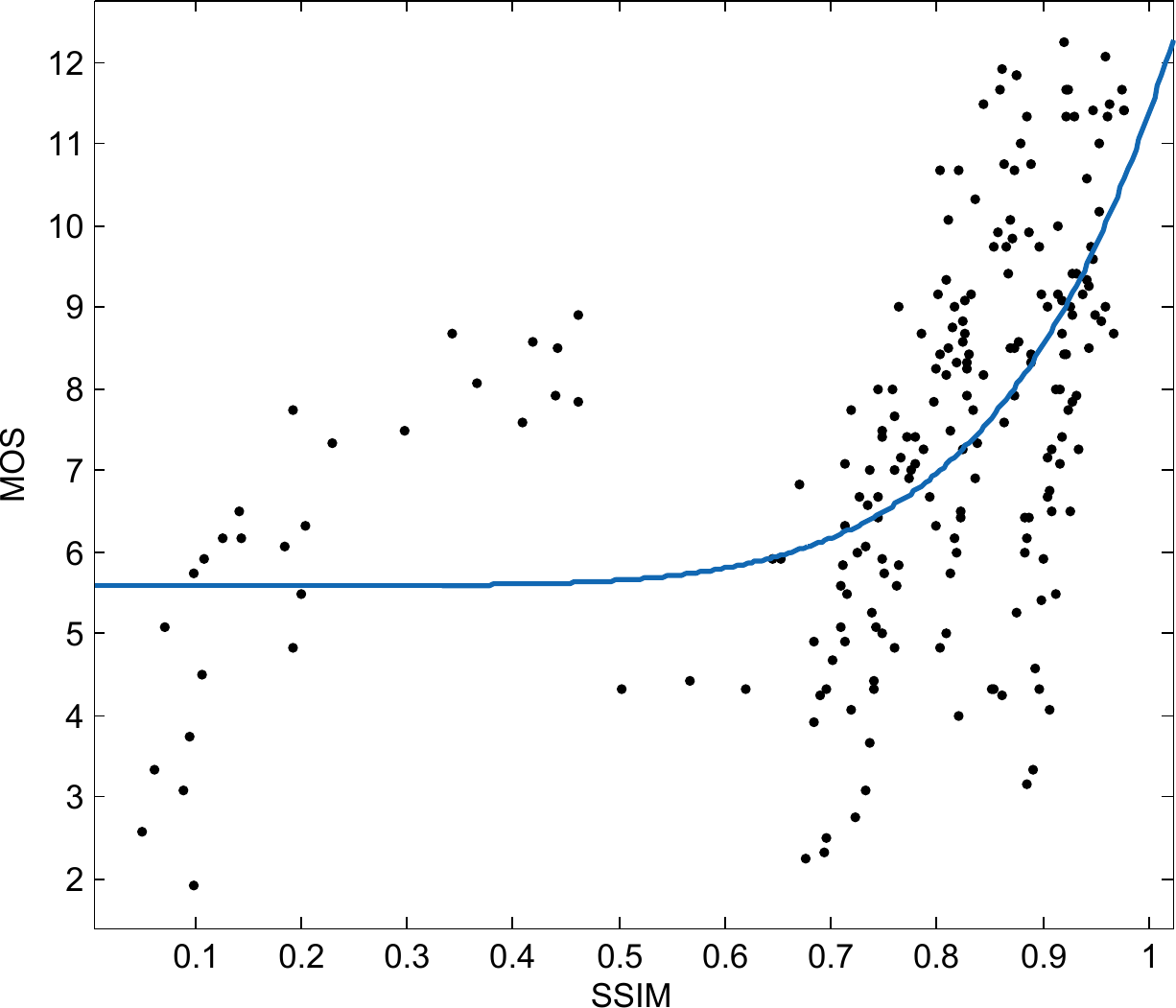}
                \caption{SSIM}
        \end{subfigure}
		\quad
        \begin{subfigure}[t]{0.175\textwidth}
                \centering
                \includegraphics[width=\textwidth]{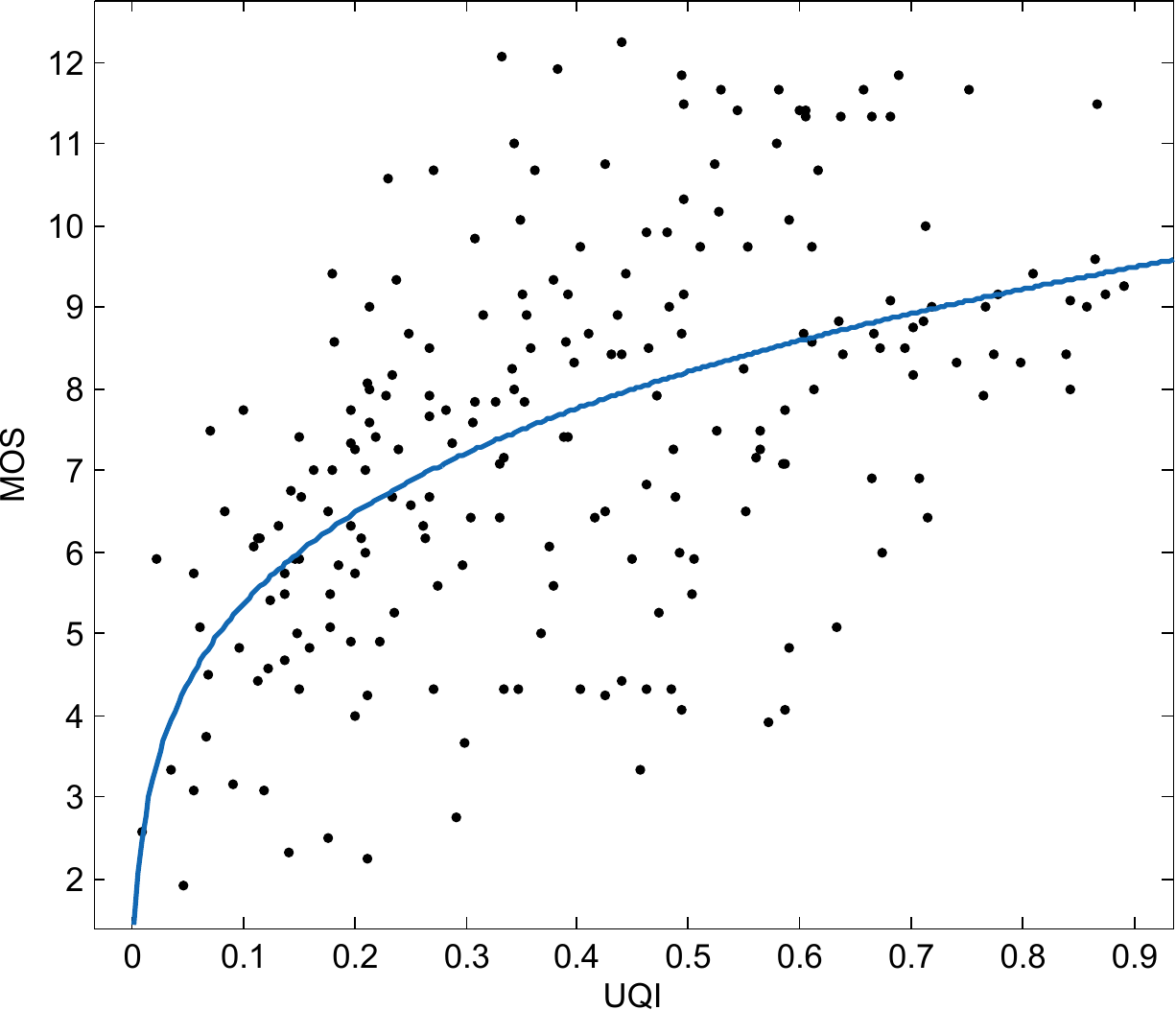}
                \caption{UQI}
        \end{subfigure}
		\quad
        \begin{subfigure}[t]{0.175\textwidth}
                \centering
                \includegraphics[width=\textwidth]{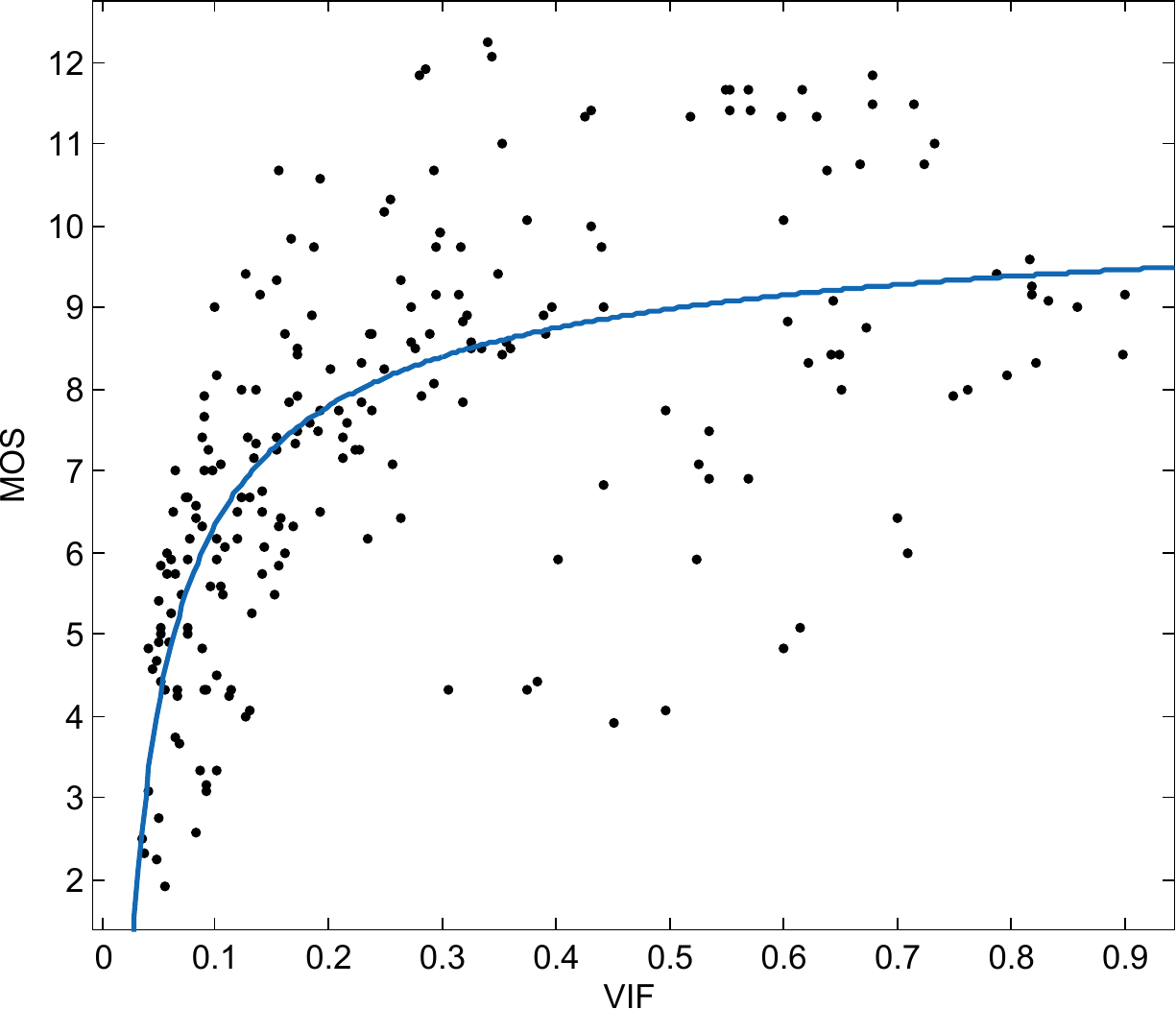}
                \caption{VIF}
        \end{subfigure}
		\quad
        \begin{subfigure}[t]{0.175\textwidth}
                \centering
                \includegraphics[width=\textwidth]{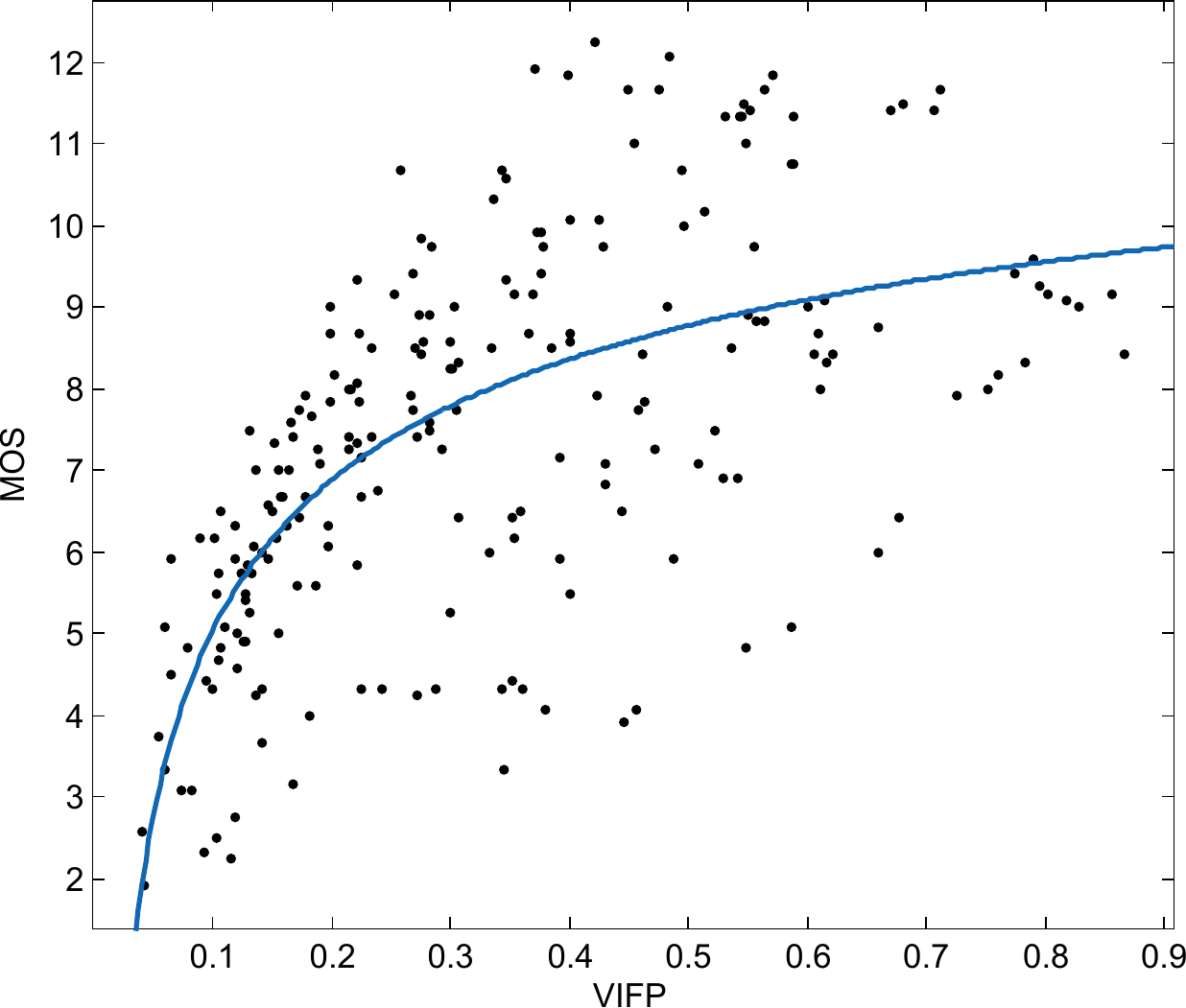}
                \caption{VIFP}
        \end{subfigure}
        \begin{subfigure}[t]{0.175\textwidth}
                \centering
                \includegraphics[width=\textwidth]{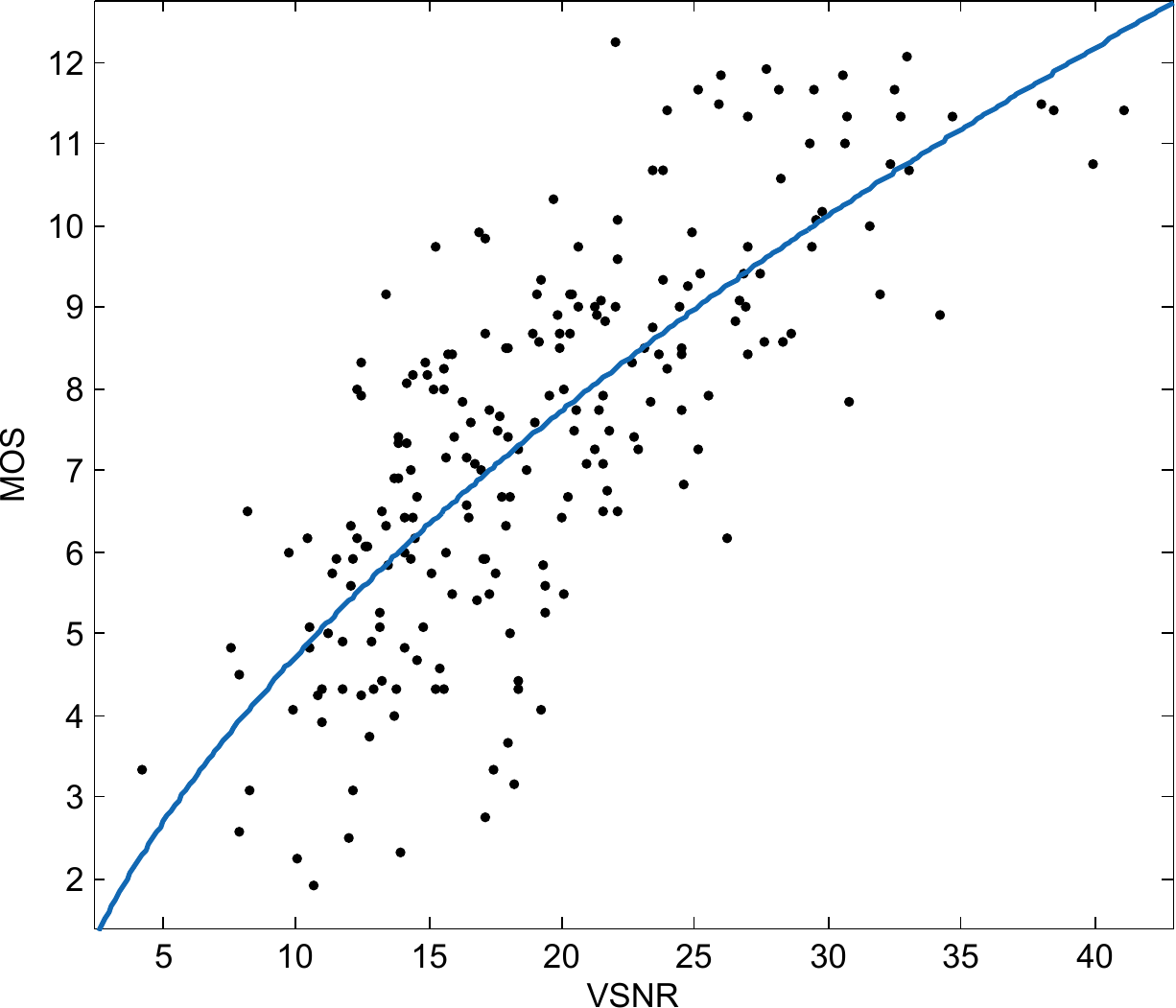}
                \caption{VSNR}
        \end{subfigure}
\caption{Scatter plots for images in Dataset C of the
MCL-3D database, where the x-axis is the subjective MOS value and the
y-axis is the predicted MOS using various objective quality
indices.}\label{fig:scatter1}
\end{figure*}

The scatter plots of predicted scores vs. MOS along
with their fitting curves approximated by function $f(x)=ax^b+c$ with
95\% confidence level for 18 objective quality metrics are shown in
Fig.~\ref{fig:scatter1}. These data are calculated based on Dataset C
of MCL-3D. Each point on the plot represents one stereoscopic image
pair. The horizontal axis gives the objective quality score while the
vertical axis indicates the MOS value. The scatter
plots confirm that the PBSIQA metric exhibits the best correlation with
human perception.  On the other hand, many of existing algorithms such
as IFC, SSIM, UQI lose the monotonic property with respect to MOS as
demonstrated in SROCC values of Table~\ref{tab:perf_comparison}.
Although the fitting curves of NQM and PSNR\_HVS show trends similar to
that PBSIQA, their variances are much larger.

\begin{table}[t]
  \centering
  \caption{Performance comparison over IVC-A and LIVE-A databases.}
    \begin{tabular}{rrcrc}
    \toprule
          & \multicolumn{4}{c}{PCC} \\
    \midrule
    Rank  & \multicolumn{2}{c}{IVC-A Database} & \multicolumn{2}{c}{LIVE-A Database} \\
    \midrule
    1     & PBSIQA & 0.952 & PBSIQA & 0.924 \\
    2     & \multicolumn{1}{c}{RKS} & 0.884 & \multicolumn{1}{c}{RKS} & 0.841 \\
    3     & \multicolumn{1}{c}{SSIM} & 0.821 & \multicolumn{1}{c}{SSIM} & 0.841 \\
    4     & \multicolumn{1}{c}{C4} & 0.811 & \multicolumn{1}{c}{Campisi} & 0.781 \\
    5     & \multicolumn{1}{c}{Campisi} & 0.746 & \multicolumn{1}{c}{C4} & 0.776 \\
    6     & \multicolumn{1}{c}{CYCLOP} & 0.711 & \multicolumn{1}{c}{PSNR} & 0.733 \\
    7     & \multicolumn{1}{c}{PSNR} & 0.671 & \multicolumn{1}{c}{VIFP} & 0.654 \\
    8     & \multicolumn{1}{c}{BQPNR} & 0.614 & \multicolumn{1}{c}{VSNR} & 0.654 \\
    9     & \multicolumn{1}{c}{UQI} & 0.591 & \multicolumn{1}{c}{CYCLOP} & 0.652 \\
    10    & \multicolumn{1}{c}{Benoit} & 0.569 & \multicolumn{1}{c}{VIF} & 0.642 \\
    11    & \multicolumn{1}{c}{VIF} & 0.568 & \multicolumn{1}{c}{BQPNR} & 0.633 \\
    12    & \multicolumn{1}{c}{NQM} & 0.555 & \multicolumn{1}{c}{UQI} & 0.631 \\
    13    & \multicolumn{1}{c}{VIFP} & 0.545 & \multicolumn{1}{c}{PSNR\_HVS} & 0.612 \\
    14    & \multicolumn{1}{c}{SNR} & 0.544 & \multicolumn{1}{c}{Benoit} & 0.607 \\
    15    & \multicolumn{1}{c}{IFC} & 0.532 & \multicolumn{1}{c}{SNR} & 0.598 \\
    16    & \multicolumn{1}{c}{PSNR\_HVS} & 0.531 & \multicolumn{1}{c}{NQM} & 0.558 \\
    17    & \multicolumn{1}{c}{MSE} & 0.523 & \multicolumn{1}{c}{MSE} & 0.546 \\
    18    & \multicolumn{1}{c}{VSNR} & 0.407 & \multicolumn{1}{c}{IFC} & 0.524 \\    
    \bottomrule
    \end{tabular}%
  \label{tab:Performance2}%
\end{table}%

\begin{table}[t]
  \centering
  \caption{Cross-data validation results by using models generated from MCL-3D to predict a quality score of stereo images in IVC-A and LIVE-A databases.}
    \begin{tabular}{>{\centering\arraybackslash}m{2cm}|>{\centering\arraybackslash}m{1.5cm}>{\centering\arraybackslash}m{1.5cm}>{\centering\arraybackslash}m{1.5cm}}
    \toprule
    \textbf{Database} & \textbf{PCC} & \textbf{SROCC} & \textbf{RMSE} \\
    \midrule
    IVC-A  & 0.884 & 0.853 & 0.404 \\
    LIVE-A  & 0.893 & 0.844 & 0.433 \\
    \bottomrule
    \end{tabular}%
  \label{tab:Performance3}%
\end{table}%

Next, we show the performance comparison over IVC-A and
LIVE-A databases in Table~\ref{tab:Performance2}. We only fuse scorers
\#1$\sim$\#7 for texture distortions in the PBSIQA system since these
two databases consist of the stereoscopic data format (i.e., the left
and the right views without the depth map). As shown in
Table~\ref{tab:Performance2}, PBSIQA still outperforms existing metrics
significantly. For example, the PCC values of PBSIQA for IVC-A and
LIVE-A are 0.952 and 0.924, respectively while those of the second best
metric (RSK) are 0.884 and 0.841. We see that the PCC values of PBSIQA
for these two databases are higher than that for MCL-3D.  This is due to
the fact that the distortion types in IVC-A and LIVE-A are simpler than
that of MCL-3D so that each scorer can generate more reliable scores for
them.

Practically, it is not convenient to train learning-based quality
assessment indices to tailor to a new database every time. It is
desirable to develop a quality metric that offers consistent performance
when it is trained by one database yet tested by a different database.
This is called cross-database validation. To conduct this task, we train
the PBSIQA system using the MCL-3D database and then use it to predict
the quality score of stereoscopic images in IVC-A and LIVE-A. The
results are shown in Table~\ref{tab:Performance3}. Although its
performance degrades slightly, the PBSIQA system still offers
good results. As the SIQA database size becomes larger, its 
prediction accuracy goes higher. Finally, it is worthwhile to point out that, 
since the PBSIQA system consists of multiple individual scorers, its 
complexity is roughly equal to the total complexity of all contributing scorers. 
Basically, it trades higher complexity for better performance.

\section{Conclusion and Future work}\label{sc:conclusions}

In this work, we proposed a ParaBoost method to design a new
stereoscopic image quality assessment system called PBSIQA, which
includes a set of parallel quality scorers designed to address various
distortions in Stage I and their scores are fused to yield the ultimate
quality score in Stage II. The excellent performance of the PBSIQA system
was collaborated by extensive experimental results.  
Our research has an impact to the 3D content delivery field since a
reliable 3D quality assessment metric serves as a valuable objective
function in optimizing the overall system performance.

As future extensions, it is interesting to incorporate the masking
effect. For example, distortions in salient regions such as foreground
objects have more negative impacts on perceptual quality. Also, textured
regions are usually more robust to distortions, whereas distortions in
homogeneous regions are more noticeable. We may further improve
performance by considering content characteristics. 

Another interesting extension is to consider the
quality assessment of 3D video. On one hand, we believe that the
ParaBoost methodology can be extended to the case of 3D video in
principle. On the other hand, we do need sufficiently large 3D video
quality assessment databases for the learning-based methodology to
apply. The lack of publicly accessible 3D video quality assessment
databases is the main barrier to this research and it is desired to
develop such databases for the research community.

\section*{Acknowledgment} Computation for the work described in this
paper was supported by the University of Southern California's Center
for High-Performance Computing (hpc.usc.edu).

\end{document}